\documentclass[3p,times]{elsarticle}
\usepackage{amssymb, amsmath}
\usepackage{graphicx}
\usepackage{url}
\usepackage{amsmath}

\usepackage{multicol}
\usepackage{epstopdf}
\usepackage{epsfig}
\usepackage{color}

\usepackage{ecrc}  
\usepackage[figuresright]{rotating}
\usepackage[T1]{fontenc}
\usepackage{amssymb}
\usepackage{amsmath}
\usepackage{amsfonts}
\usepackage{verbatim}

\usepackage{hyperref}
\usepackage{array}
\usepackage{algorithm,algpseudocode}
\usepackage{placeins}
\usepackage{xcolor,colortbl}

\definecolor{Gray}{gray}{0.85}
\newcolumntype{a}{>{\columncolor{Gray}}c}

\volume{00}

\firstpage{1}

\journalname{Procedia Computer Science}

\runauth{}

\jid{procs}

\jnltitlelogo{Procedia Computer Science}

\begin{document}


\begin{frontmatter}

\dochead{}

%
\title{An efficient FPGA implementation of Anisotropic diffusion filtering on Images}

\author{Chandrajit Pal$^1$, Amlan Chakrabarti$^2$, 
and Ranjan Ghosh$^3$
}

\address{Corresponding author mail:palchandrajit@gmail.com 1.
}

\address{A.K.Choudhury School of Information Technology1,2, 
Institute of Radio Physics and Electronics3, 
University of Calcutta, 92 A. P. C. Road, Kolkata:700 009, India.
}

\title{A Brief Survey of Recent Edge-Preserving
Smoothing Algorithms on Digital Images
}

%

%
%


\begin{abstract}

Edge preserving filters preserve the edges and its information while blurring an image. In other words they are used to smooth an image, while reducing the edge blurring effects across the edge like halos, phantom etc. They are nonlinear in nature.
Examples are bilateral filter, anisotropic diffusion filter, guided filter, trilateral filter etc.
Hence these family of filters are very useful in reducing the noise in an image making it very demanding in computer vision and computational photography applications like denoising, video abstraction, demosaicing, optical-flow estimation, stereo  matching, tone mapping, style transfer, relighting etc.
 This paper provides a concrete introduction to edge preserving filters starting from the heat diffusion equation in olden to recent eras, an overview of its numerous applications, as well as mathematical analysis, various efficient and optimized ways of implementation and their interrelationships, keeping focus on preserving the boundaries, spikes and canyons in presence of noise. Furthermore it provides a realistic notion for efficient implementation with a research scope for hardware realization for further acceleration.\\


\end{abstract}

\end{frontmatter}

\section{Introduction}

From the very beginning there a rapid increasing interest in models and methods for discontinuous
phenomena, both in the computer graphics and vision community. The main focus lies in the identification of discontinuities in data perturbed by noise. To be more specific, in image data where noise needs to be removed from image while preserving basic significant features like gradients, jumps, spikes, edges and boundaries.

 Methods dealing with region boundary preservation have a long tradition in imaging and in their state of self applied art. Each one have their promising contributions from various branches of mathematical perspective in their respective application domain.

An evolutionary review study has been performed with detail analysis of every method starting from diffusion process{\cite{perona}} , nonlinear Gaussian filters{\cite{nonlinear}}, bilateral filter{\cite{tomasi,porikli,chen,kunalda}} to its extension the trilateral filter {\cite{prasun,vaudrey}}.

Edge preserving filtering is a technique to smooth images while preserving edges. It all started in 1985 with the work of L.P  Yaroslavsky \cite{yaroslavsky}, with his neighborhood filter which averages pixels with a similar grey level value and belong-
ing to the spatial neighborhood.  In 1990  Perona and Malik \cite{perona} worked on  realizing semantically meaningful edges using a diffusion process. To overcome some of its limitation Aurich and Weule in the year 1995 made a breakthrough invention on edge preserving method using non-linear Gaussian filters. Their concept laid the foundation stone of the origin of bilateral filters.  It was later rediscovered by Smith and Brady \cite{smith} , and Tomasi
and Manduchi \cite{tomasi} in the year 1998 gave it its current name 'the Bilateral filter'. Ever since it has found a widespread use in various image processing applications like video abstraction, demosaicing, optical-flow estimation, stereo  matching,  tone mapping, stylization, relighting and denoising. Its various advantages lies firstly with its non-iterative and simple formulation where each pixel is replaced by a weighted average of its neighborhood pixels making it adaptive to application specific ambiance. Secondly it depends on few parameters (spatial domain and intensity range kernels). Finally with the superb contributions of various efficient numerical schemes the filter can be computed very fast reasonably on large images also in real time environment with dedicated graphics hardware.

A great deal of implemented papers \cite{tomasi,porikli,chen,pham,dynamic,kunalda,cviu} have been studied, which explores the various feature of the edge preserving filters(mainly bilateral filter) constituting its behaviour and gives a detail understanding of its strengths and weaknesses in various application environment domains, the various optimization techniques for further betterment which consequently lead to its extension the next generation trilateral filter \cite{prasun,vaudrey}. It provides stronger noise reduction and better outlier rejection in high-gradient regions, and it mimics the edge-limited smoothing behavior of shock-forming PDEs by region finding with a fast min-max stack. This study is supposed to provide a good insight to the designers among  the selection of the design methods of their choice deciding upon the various numerical analysis compatible with their applications in hand. 

This paper is organized as follows. Section 2 presents the diffusion technique employed to detect the edge, namely the anisotropic diffusion, next realizing the same through nonlinear Gaussian filters,  linear Gaussian filtering and the nonlinear extension to the bilateral filter and determining the computational cost, number of iterations followed by the various effects on it by changing its parameters. Section 3 discusses the various ways to efficiently implement the bilateral filter together with the variations of each method to achieve the other and also the several links mainly of the bilateral filter to other frameworks employing the different methods to interpret it. Section 4 describes various challenging applications of the edge-preserving filters. Section 5 describes the various interrelationships among the edge preserving filters with an in depth mathematical analysis of each transition. Section 6 exhibits the various modifications, extensions and substitutes of the bilateral filter to the next generation trilateral filters.

\section{Edge preserving through nonlinear diffusion to trilateral filtering}

To introduce the concept of edge preserving we begin with our discussion with the nonlinear diffusion technique by Perona and Malik \cite{perona}. They proposed a nonlinear diffusion method for avoiding the blurring and localization problems caused by linear diffusion filtering. It aims at reducing image noise without removing significant parts of the image content, typically lines, edges or other details that are important for the interpretation of the image.
\subsection{Motivation leading to realizing the anisotropic diffusion filter}

Suppose we have an edge/not edge estimation function
$E$ such that 

\begin{equation}
\textit{E}(x,y,t)=\bigtriangledown I(x,y,t) 
\label{eq:def1}
\end{equation}
 where

\begin{equation}
\bigtriangledown I(x,y,t)=(\frac{\partial I }{\partial x},\frac{\partial I}{\partial y})
\label{eq:def2}
\end{equation}


%


 If (x,y) is a part of an edge then naturally little smoothing should be applied unlike otherwise. Fortunately the gradient of the brightness function acts as a detector which tells whether (x,y) is a part of an edge or not denoted by $\bigtriangledown I(x,y,t)=(\frac{\partial }{\partial x},\frac{\partial }{\partial y})$. Let $\Omega \subset\mathbb{R}^{2} $ denote a subset of the plane and \textit{I}(.,t):$\Omega\rightarrow \mathbb{R}$ be a family of gray scale images therefore anisotropic diffusion is defined as 

\begin{equation}
\frac{\partial I}{\partial t}=\textit{div}(c(\textit(x,y,t)\Delta I))
=\bigtriangledown c \bigtriangledown I+c(x,y,t)\Delta I
\end{equation}
where $\Delta $ and $\bigtriangledown$ denotes the laplacian and gradient respectively, $\textit{div(...)}$ is the divergence operator and $c(x,y,t)$ is the diffusion coefficient, $c(x,y,t)=g(\left \|E(x,y,t) \right\|)$ controls the rate of diffusion i.e blurring intensity according to $\left \|E(x,y,t) \right\|$ and is the function of the image gradient so as to preserve the edges in the image. When $c(x,y,t)$ is large (x,y) is not a part of an edge else otherwise. The edge estimation function is E(x,y,t)=$\bigtriangledown I(x,y,t)$
 correspondingly the two functions for the diffusion coefficient, the corresponding two functions for the diffusion coefficients are 
\begin{equation}
\ c(\left \| E \right \| )=e^-({\left \|E \right\|/k})^{2}
 and $$
 $$\ c(\left \| E \right \| )=1/(1+{\left \|E \right\|/k})^{2}
\end{equation}
where the constant $k$ controls the sensitivity to the edges and is either chosen experimentally or as a function of the image noise. \\ \\ 
\begin{figure}
\centering
\includegraphics[height=5cm]{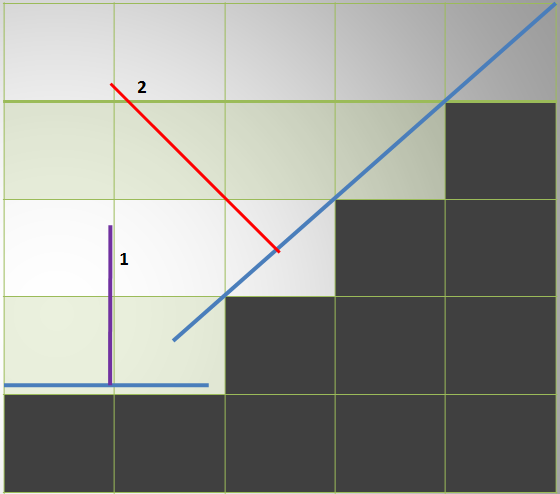}
\caption{In the first part(1),
the gradient vector is (0, n)     while (n$ >$0).
In the second part(2), the gradient vector is (-m , m) 
while  (m $>$0 ). These points where the norm of the gradient is high, could be treated as edge points, end therefore be applied less blurring.
}
\label{fig:fig1}
\end{figure}

If gradient is (a,b), then a perpendicular vector to it, can be (a, -b). A dynamic kernel is contracted along the direction of the normal, ending in an elliptical kernel. \\ We have successfully implemented this algorithm after identifying the parallelism inherent with it on reconfigurable hardware and have attained a prominent acceleration \cite{elsevier}. We have presented the FPGA implementation of an edge preserving anisotropic diffusion filter for digital images. The designed architecture completely replaced the convolution operation and realized the same using simple arithmetic subtraction of the neighboring intensities within a kernel, preceded by multiple operations in parallel within the kernel. Our proposed hardware design architecture completely substituted standard convolution operation \cite{gonzaleg}, required for evaluating the intensity gradient within the mask. We used simple arithmetic subtraction to calculate the intensity gradients of the neighboring pixels, which saved 9 multiplication and 8 addition operations per convolution respectively by computing only a single arithmetic operation for each gradient direction as per equation 5. Thus, reducing a huge amount of computational time and operations to a linear computational complexity. The hardware implementation of the proposed design shows better performance in terms of throughput ( increase of 10\% ), total power ( reduction of 24\% ) and a reduced resource utilization in comparison with the related research works In the later sections we will show the relationship of this algorithm with its counterparts.

\begin{figure}
\centering
\includegraphics[height=4.2cm]{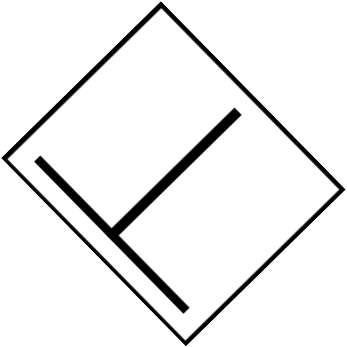}
\caption{ The  desired \emph{direction} of the dynamic kernel is effected by the local $gradient’s$ direction $\bigtriangledown I(x,y,t)$ and it is perpendicular to it.
}
\label{fig:fig2}
\end{figure}

\begin{figure}
\centering
\includegraphics[height=4.2cm]{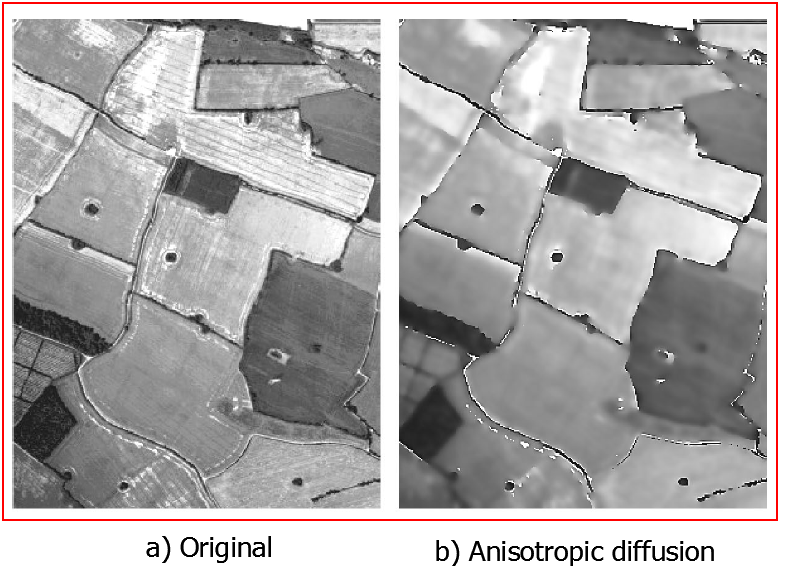}
\caption{ fig 'a' denotes the original figure and 'b' denotes the denoised image after applying anisotropic diffusion i.e Perona Malik filter.
}
\label{fig:fig3}
\end{figure}

\subsection{Edge preserving diffusion through Non-Linear Gaussian filters}

Aurich and Weule in the year 1995 \cite{nonlinear} proposed a new diffusion method for edge preserving smoothing of images. Unlike the anisotropic diffusion methodology it is based on a modification of the way the solution of the heat conductance equation is obtained by convolving the Gaussian kernel with the initial data. Their method employs a simple non-linear modifications of the Gaussian filters thereby overcoming the lengthy iteration steps and convergence problems. A sequential chain of non-linear filters combining the significant as well as advantageous aspects of both the anisotropic diffusion and robust filtering has been applied overcoming their drawbacks.

\subsubsection{The non-linear Gaussian filtering approach} 

A simple linear Gaussian filtering convolution is defined by Gf=\textit{g*f } where \textit{f} is the signal and \textit{g} is the Gaussian function. The image signals are defined on a discrete bounded set $\rho $ of pixels and as a result region boundary effects needs to be taken care off.
The linear Gaussian filter using the position-dependent renormalization of the weights is defined by

\begin{eqnarray}
G_{\sigma_{ x}}\textit{f(p)}=\frac{1}{N_{p}}\sum_{q\epsilon \rho}^{} \ g_{\sigma_{ x}}(||q-p ||)f(q)   \\
 =f(p) + \frac{1}{N_{p}}\sum_{q\epsilon \rho}^{} \ g_{\sigma_{ x}}(||q-p ||)(f(q)-f(p))  \nonumber 
\end{eqnarray}
  with   $g_{\sigma_{x}}(t)$=$exp({{-t^{2}}/2\sigma_{x}^{2}})$ and  $N_{p}=$$\sum_{q\epsilon \rho}^{} \ g_{\sigma_{ x}}(||q-p ||).$ 
            
In order to preserve edges one should avoid averaging across edges. Thus if q
and p are at different sides of an edge \textit{f(q)} should be disregarded while
calculating the average over a neigbourhood of \textit{p}. Usually, however, the location
of the edges is not yet known, even worse: Edge-preserving is done in order to
detect edges more easily. Therefore the value \textit{f(q)} is not completey omitted in
the averaging process, but its weight is lowered when it is likely that there is
an edge between q and p. How to estimate this probability? The
difference $\left | \textit{f(q)-f(p)} \right |$ needs to be looked upon. The bigger it is, the more probable it seems that there is an edge between \textit{p} and \textit{q}. This has been realised by multiplying $g_{\sigma_{ x}}(||q-p||)$ by an additional weight $\psi({f(q)-f(p)}).$  As because the noise distribution in images in real situations is bell-shaped i.e Gaussian like so it is suggestive to choose $\psi=g_{\sigma_{x}}$ i.e also as Gaussian, thereby generating the non-linear Gaussian filter given by 
\begin{equation}
G_{\sigma_{x},\sigma_{z}}=f(p) +\frac{1}{N_{p}}\sum_{q\epsilon \rho}^{} \ g_{\sigma_{ x}}(||q-p ||)g_{\sigma_{ z}}(f(q)-f(p)).(f(q)-f(p)).  \label{eq:xdef8}
 \end{equation}
where
\begin{equation}
N_{p}=\sum_{q\epsilon \rho}^{} \ g_{\sigma_{ x}}(||q-p ||)g_{\sigma_{ z}}(f(q)-f(p)). \label{eq:xdef9}
 \end{equation}

It is well known that the Gaussian filter is the basic solution of the heat conductance equation, therefore the non-linear Gaussian filter can be utilized to calculate some form of anisotropic diffusion.
With an aim to further decrease the noise an extra parameter $\eta$ has been added to the nonlinear Gaussian filter equation \ref{eq:xdef9} resulting in 

\begin{equation}
 G_{\sigma_{x},\sigma_{z},\eta}(p)=f(p)+\frac{\eta}{N_{p}}\sum_{q\epsilon \rho}^{}\ g_{\sigma_{ x}}(||q-p ||)g_{\sigma_{ z}}(f(q)-f(p)).(f(q)-f(p)).
\end{equation}
Experimental observations have revealed that larger the value of $\eta$ narrower is the noise distribution of the filters as shown in fig 4 below.

\begin{figure}
\centering
\includegraphics[height=6.7cm]{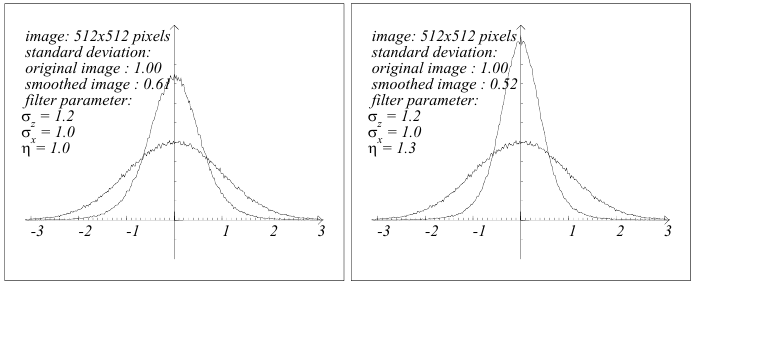}
\caption{ shows the effect of the parameter $\eta$ with 1.0 and 1.3 at the left and right figures respectively. Courtesy of \cite{nonlinear}.
}
\label{fig:fig4}
\end{figure} 

The value of the parameters namely $\sigma_{x}$ and $\sigma_{z}$ plays an important role in determining the shape of the output signal when the filter is applied on step and ramp like edges. \\
Case 1: When $\sigma_{z}$ is larger than the maximum grey value the region boundaries tend to smooth as non-linear Gaussian filter tends to work like a linear Gaussian filter. \\
Case 2: When $\sigma_{x}$ is large and $\sigma_{z}$ is small the filter can sharpen ramp edges as shown in fig 5 below.

\begin{figure}
\centering
\includegraphics[height=2.7cm]{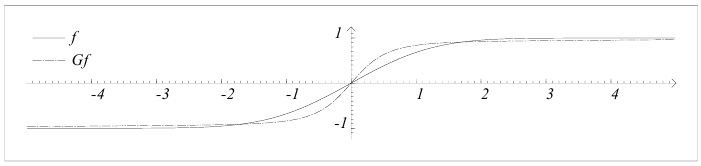}
\caption{ shows the effect of sharpening of a blurred edge. Courtesy of \cite{nonlinear}.
}
\label{fig:fig5}
\end{figure} 

It has been noticed that $\sigma_{x}$ should be chosen greater than 1 to sharpen the ramp edge \textit{f}.

According to the authors \cite{nonlinear} a sequence of filtering operations are applied with varying parameters in order to achieve the desired smoothing effects. The first filtering step aims to reduce the contrast of the fine details in the image followed by the subsequent steps which does the same thing but tries to sharpen the edges of the coarser structures which is supposed to be blurred by the first filtering step.

\subsubsection{Effect of varying parameters} In the first step $\sigma_{x}$ should be very small so as to reduce the unwanted blurring of the coarser structures but at least larger than the size of the fine structures. The $\sigma_{z}$ on the other hand determines the maximal contrast fine details approximately in order to be finally eliminated as a result. 
From the next filtering steps the $\sigma_{x}$ needs to be increased and $\sigma_{z}$ to be decreased to sharpen the edges of the coarser structure. Experiments have proved that almost always best results are achieved if $\sigma_{x}$ is doubled and $\sigma_{z}$ is halved before performing the next filtering step. 

\subsubsection{Results} Fig 6 below shows the result of filtering with a filter chain of five stages with initial parameters as $\sigma_{x}$=0.5 and $\sigma_{z}$=128 respectively.

\begin{figure}
\centering
\includegraphics[height=4cm]{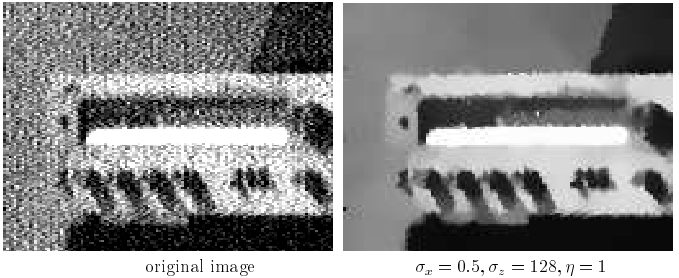}
\caption{Smoothing the picture of a radio with five 5 filtering stages with $\sigma_{x}$=0.5 and $\sigma_{z}$=128. Figure reproduced from \cite{nonlinear}.
}
\label{fig:fig6}
\end{figure}

This design introduces some flexibility w.r.t its anisotropic counterpart as this Gaussian filtering can be considered as an anisotropic diffusion process which can be stopped at certain times and can be restarted later with new parameters.

\begin{equation}
\frac{1}{N_{p}}\sum_{q\epsilon \rho}^{} \ g_{\sigma_{ x}}(||q-p||)g_{\sigma_{ z}}(f_{o}(q)-f_{o}(p)).(f(q)-f(p)).
 \end{equation}
where
\begin{equation}
N_{p}=\sum_{q\epsilon \rho}^{} \ g_{\sigma_{ x}}(||q-p ||)g_{\sigma_{ z}}(f_{o}(q)-f_{o}(p)).
 \end{equation}

This operator smooths \textit{f} while retaining the edges present in $\textit{f}_{o}$.

\subsection{Nonlinear Gaussian filtering extension to bilateral filtering}

\subsubsection{Gaussian smoothing to Bilateral filtering}
A Gaussian blur (a.k.a Gaussian smoothing) is the result of blurring an image by a Gaussian function typically used to reduce image noise and reduce detail. Each output image pixel value is a weighted sum of its neighbors in the input image. Mathematically, applying a Gaussian blur to an image is nothing but convolving the image with a Gaussian function.\\
An image filtered by Gaussian Convolution is given by:

\begin{equation}
G[I]_{p} = \sum_{q\in S}^{} G_{\sigma}(\left\Vert p-q \right\Vert)I_{q}
\end{equation}
where
\begin{equation}
G_{\sigma }(x)= 1/2\pi\sigma^{2}exp(-(x^{2}+y^{2})/2\sigma^{2})
\end{equation}
denotes the 2D Gaussian kernel. It is a weighted average of the intensity of the
adjacent positions with a weight decreasing with the spatial distance to the center position \emph{p}. The weight for pixel \emph{q} is defined by the Gaussian 
\begin{equation}
		G_{\sigma }=(\left\Vert  p-q\right\Vert)
\end{equation}
where $\sigma$ is a parameter defining the neighborhood size. \\
\subsubsection{Drawbacks of Gaussian smoothing}
Suppose a bright pixel has a strong influence over an adjacent dark pixel although these two pixel intensity values are quite different. As a result, after convolving image edges are blurred because pixels across discontinuities are averaged together. Which concludes that action of the Gaussian convolution is independent of the image content but depends only their distance in the image.\\

\begin{figure}
\centering
\includegraphics[height=8cm, width=15 cm]{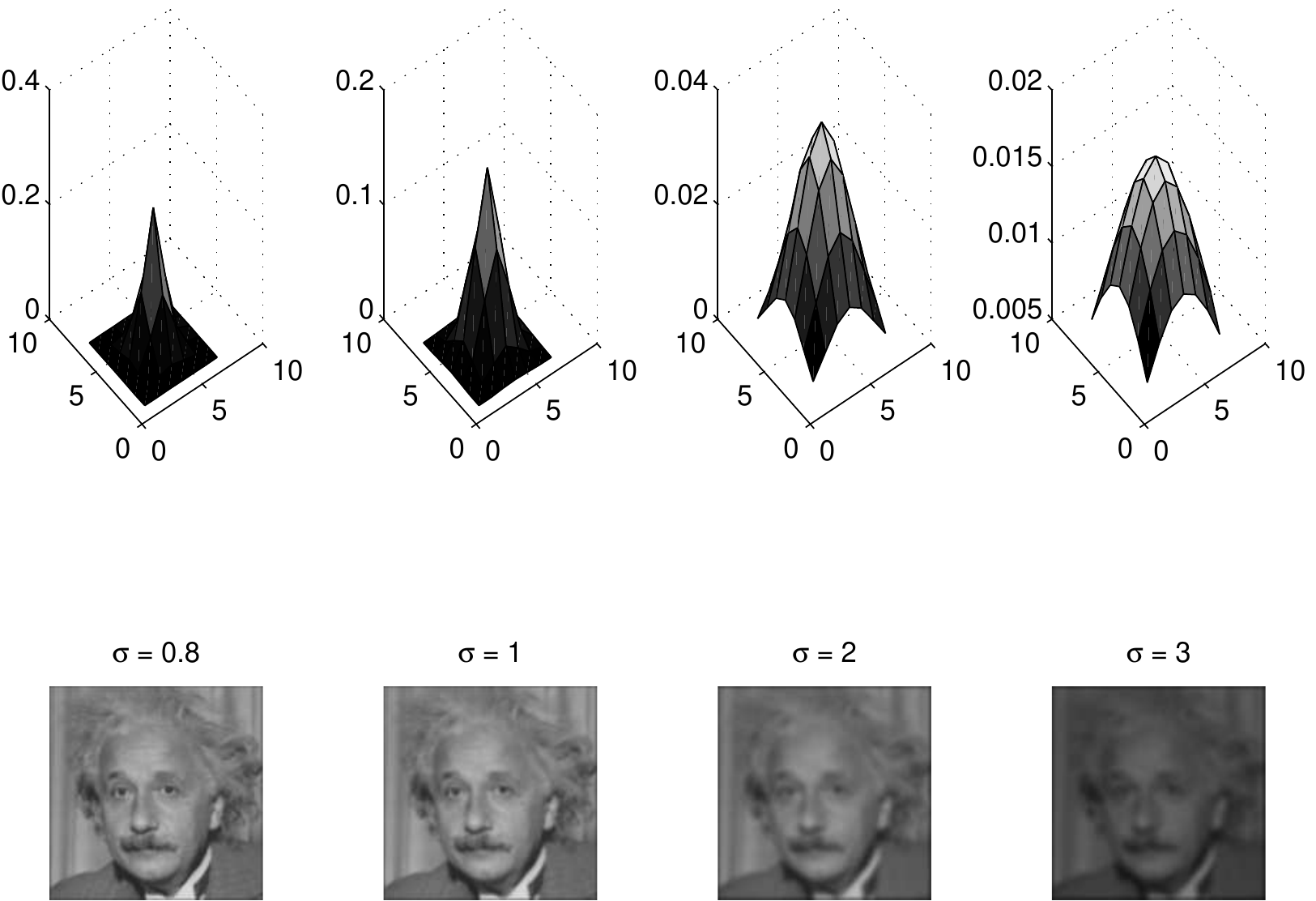}
\caption{Example of Gaussian linear filtering with varying $\sigma$. Top row shows the profile of a 2D Gaussian kernel and bottom row the result obtained by the corresponding 2D Gaussian convolution filtering. Edges are not preserved with high values of $\sigma$ because averaging is performed over a much larger area.
}
\label{fig:gauss}
\end{figure}

What is a bilateral filter? \\

Before defining it at first the concept of domain and range filtering should be discussed, origin of which is linked to the nonlinear Gaussian filter as discussed above. Traditional filtering is domain filtering which focuses on the closeness by weighing pixel values with coefficients which fall of with distance. Similarly range filtering averages image values with weight that decrease with intensity dissimilarity. It is this component which preserves the edges while averaging the homogeneous intensity regions.
 More likely to mention is that range filtering is non linear because their weights depend upon the neighborhood image intensity or color. Spatial locality is itself very promising and range filtering itself sometimes distorts an image's color map. So when the domain as well as range filtering is combined together with both the advantages the duo is termed as bilateral filtering and is defined by,\\
\begin{eqnarray}
BF[I]_{p}=\frac{1}{W_{p}}\sum_{q\epsilon \rho}^{} \ g_{\sigma_{ s}}(||q-p ||)g_{\sigma_{ r}}(f(q)-f(p)).(f(q)). where
\end{eqnarray}
\begin{equation}
W_{p}=\sum_{q\epsilon \rho}^{} \ g_{\sigma_{ s}}(||
q-p ||)g_{\sigma_{ r}}(f(q)-f(p)).
 \end{equation}

where $W_{p}$ is the normalization factor, $\sigma_{s}$ and $\sigma_{r}$ denotes the standard deviation for the domain and range kernel, accordingly $G_{\sigma_{s}}$  denotes the spatial Gaussian weighting whose value decreases with increase in pixel distance and $G_{\sigma_{r}}$ which decreases with the influence of intensity difference i.e $|f(q)-f(p)|$ between pixels say \textit{p} and \textit{q}. 
 Here we need to figure out the original framework from where the bilateral filter equation has been derived. So we need to look back in equation \ref{eq:xdef8} and equation \ref{eq:xdef9} where see that the additional term $\psi({f(q)-f(p)})$ that has been added in equation 8 to reduce the influence of \textit{f(q)} as $|f(q)-f(p)|$ increases and later on  $\psi=g_{\sigma_{z}}$ is assigned. So this $\sigma_{z}$ is actually the basis of the similarity function whose value is high in homogeneous regions within a boundary and drastically reduces across boundary on opposite sides of an edge and is analogous to the $g_{\sigma_{r}}$ in equation 13 which is nothing but the range kernel which is responsible for the sensitiveness to edges.

 \begin{figure}
\centering
\includegraphics[height=3cm, width=16 cm]{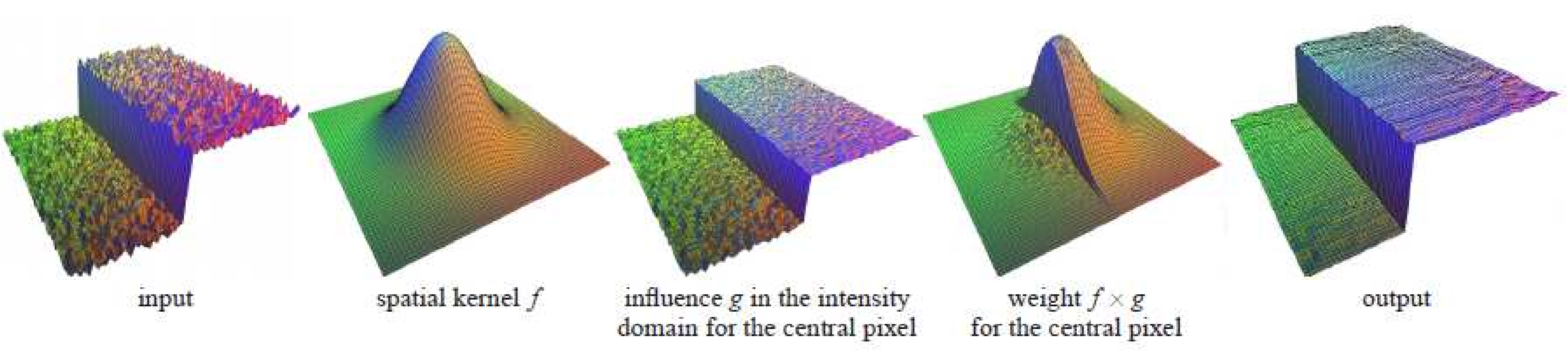}
\caption{The bilateral filters. Each pixel is replaced by a weighted average of its neighbors. Each neighbor is weighted by a spatial component that penalizes distant pixels and range component that penalizes pixels with a different intensity. Their combination ensures that only nearby similar pixels contribute to the final output. 
Courtesy of \cite{dynamic}.}
\label{fig:filtergraph}
\end{figure} 

\subsubsection{Filter controlling parameters}

Unlike anisotropic diffusion filtering, the bilateral filter is controlled by only two parameters to control its behavior namely the domain($\sigma_{s}$) and range kernel($\sigma_{r}$) standard deviations respectively. 

\textbf{\emph{Range kernel std. dev.($\sigma_{r}$)}}.
With the increase in $\sigma_{r}$ the filter gradually approximates towards the normal Gaussian convolution. As the range component increases it flattens and widens resulting in an increase in dynamic range of the image intensity so its very likely to penalize the identification of intensity gradient present at the edges.

\textbf{\emph{Domain kernel std. dev.($\sigma_{s}$)}}.
The effect of variation of domain kernel std. dev. as already been shown in fig. \ref{fig:gauss}. It smooths the large features and blurs the detailed textures upon increasing. The response of the bilateral filter is the multiplication of its constituent domain and range kernel respectively. For a large spatial Gaussian multiplied with narrow range Gaussian achieves limited smoothing in spite of the large spatial extent. The range weight enforces a strict preservation of the gradient contours.

\subsubsection{Iteration effect}

\begin{figure}
\centering
\includegraphics[height=9cm, width=16 cm]{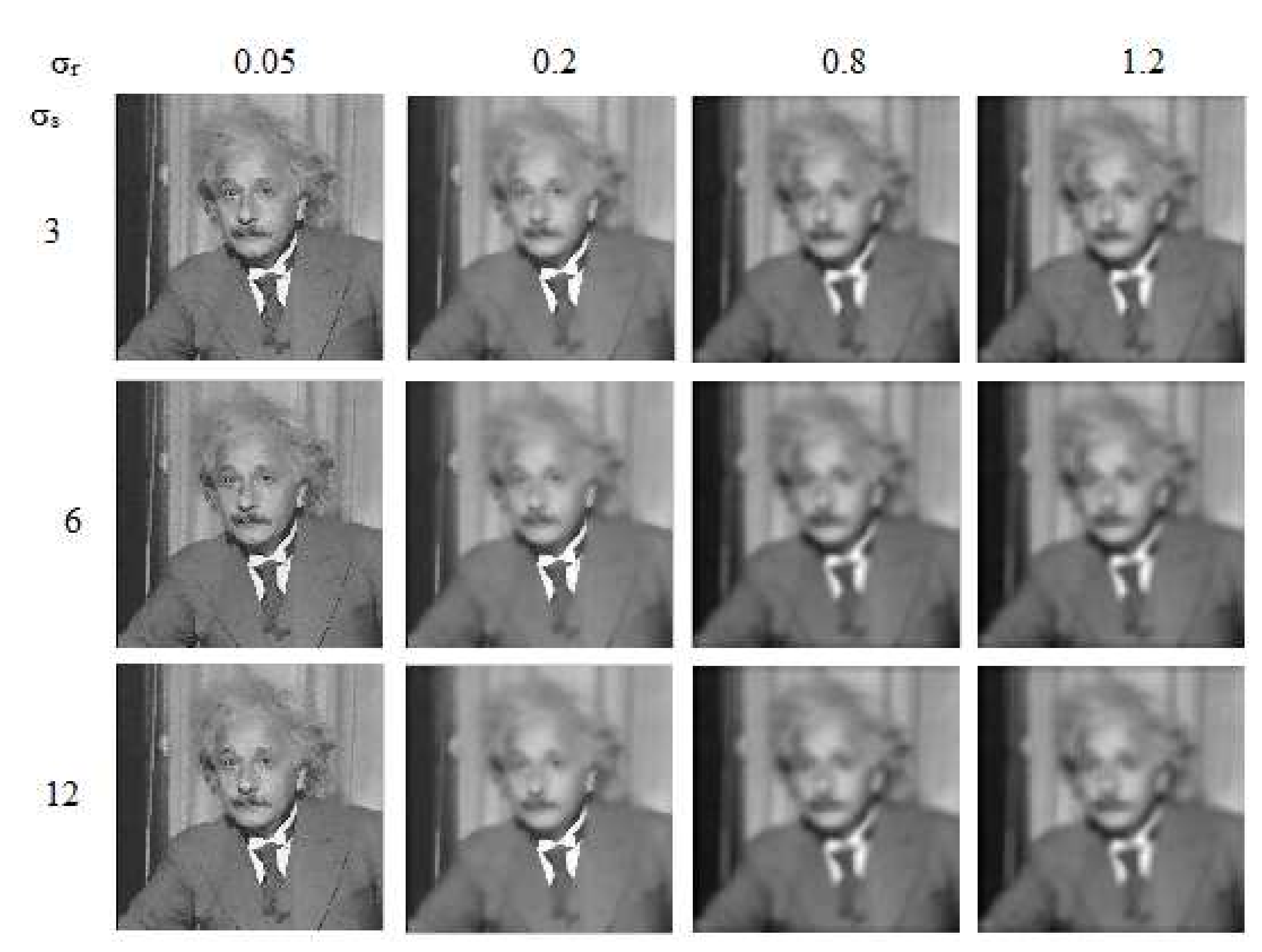}
\caption{The bilateral filters. Each pixel is replaced by a weighted average of its neighbors. Each neighbor is weighted by a spatial component that penalizes distant pixels and range component that penalizes pixels with a different intensity. Their combination ensures that only nearby similar pixels contribute to the final output. 
Courtesy of \cite{dynamic}.}
\label{fig:sigmasr}
\end{figure}

Bilateral filter is inherently non-iterative in nature. However it can be iterated. Iteration leads to almost piecewise constant result as shown in fig. \ref{fig:iteration}. Likewise we try to relate the effect of iteration with the varying filter parameters like $\sigma_r$ and $\sigma_s$. It has been observed that since $\sigma_r$ and $\sigma_s$ are interlinked the effect of varying one doesn't reflect much without the other parameter as shown in fig. \ref{fig:sigmasr}.

\subsubsection{Computational complexity}

There is no doubt that the bilateral filter is a very cost effective algorithm to compute specially when the $\sigma_s$ is large as it needs to compute a huge neighborhood of pixels, compute the weights of two parameters $\sigma_s$ and $\sigma_r$, followed by their product and an expensive normalization step.

\subsubsection{Decomposition into multiple components}

As shown in fig. \ref{fig:separation} the bilateral filter can decompose an image into two parts namely the filtered image and the subtracted/residual component. The filtered image contains the large scale features, sacrificing the minute textures respecting the strong edges. The remnant image left after subtracting the filtered image with the original one, contains only those portions the filter removed, which can be treated further as either as fine textures or noise as the requirement demands.     
 So we see that bilateral filter provides an efficient way to smooth an image while preserving the edges and has got various efficient numerical schemes to realize the same together with its wide variety of application areas to be discussed in detail in the next subsequent sections.

\begin{figure}
\centering
\includegraphics[height=4 cm, width=12 cm]{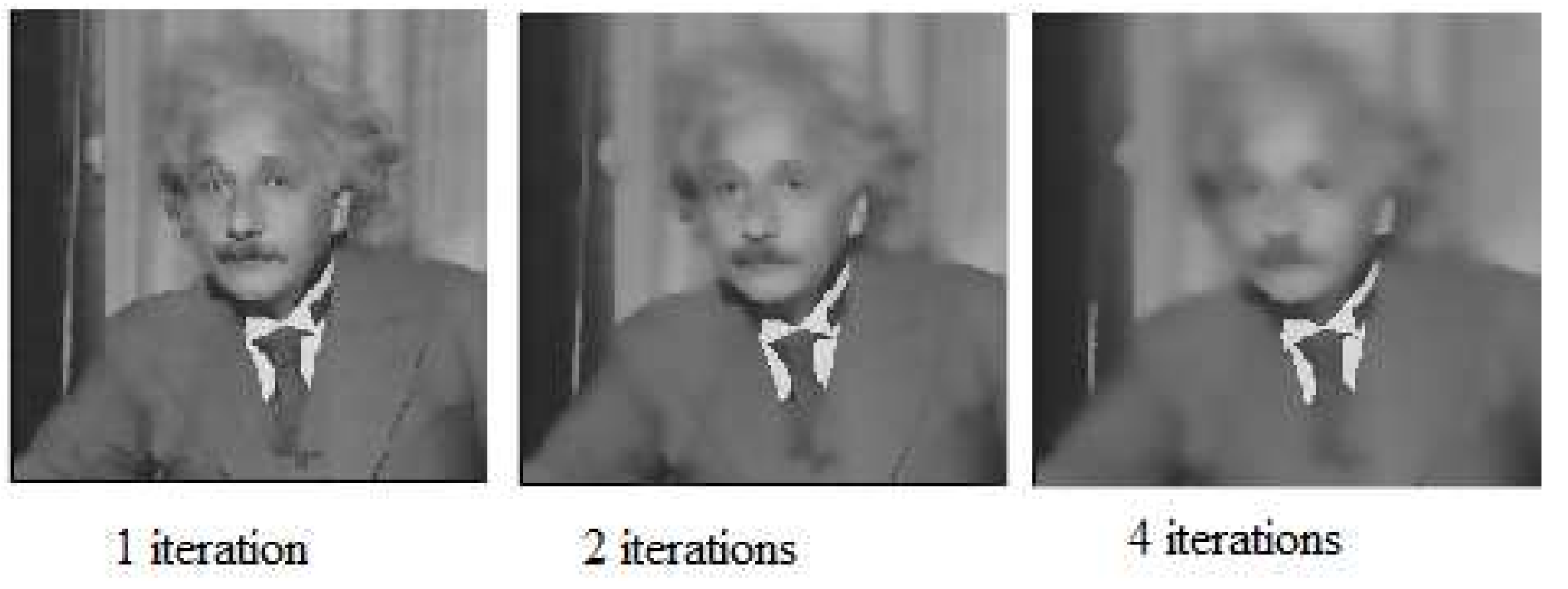}
\caption{The bilateral filters. Each pixel is replaced by a weighted average of its neighbors. Each neighbor is weighted by a spatial component that penalizes distant pixels and range component that penalizes pixels with a different intensity. Their combination ensures that only nearby similar pixels contribute to the final output. 
Courtesy of \cite{dynamic}.}
\label{fig:iteration}
\end{figure}

 \begin{figure}
\centering
\includegraphics[height=4cm, width=12.5 cm]{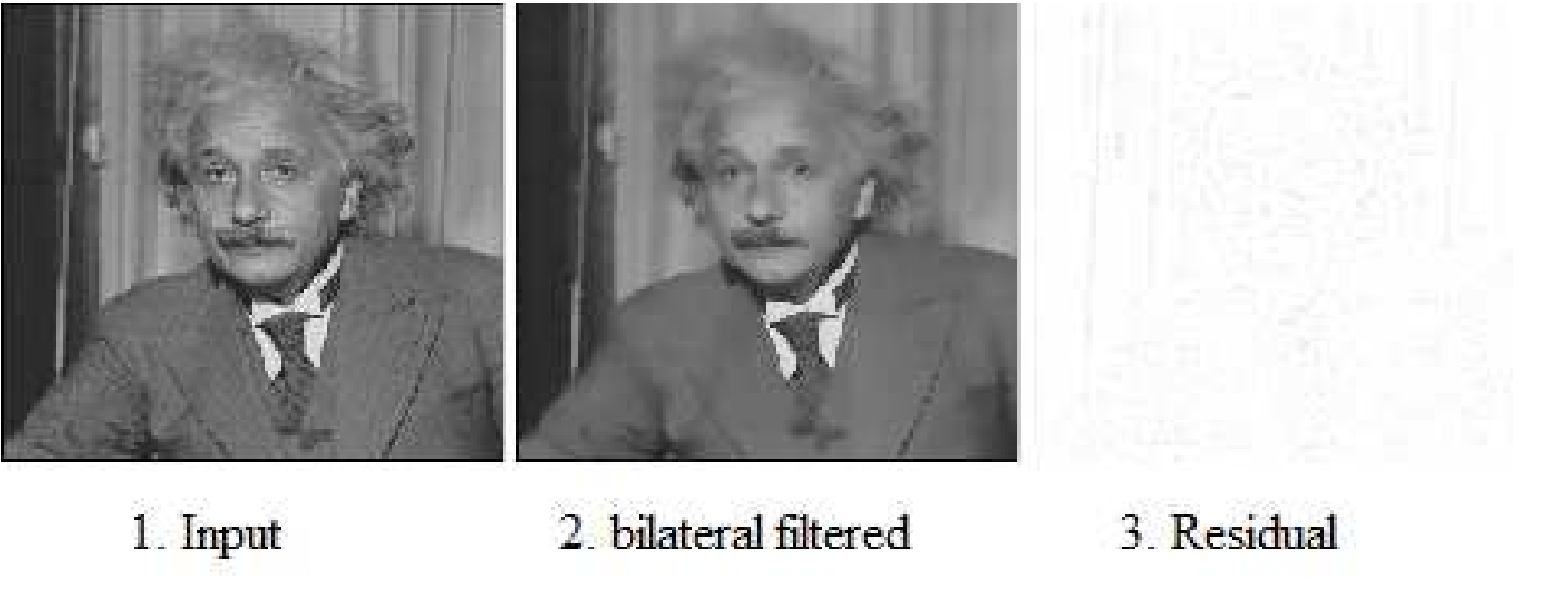}
\caption{The bilateral filters. Each pixel is replaced by a weighted average of its neighbors. Each neighbor is weighted by a spatial component that penalizes distant pixels and range component that penalizes pixels with a different intensity. Their combination ensures that only nearby similar pixels contribute to the final output. 
Courtesy of \cite{dynamic}.}
\label{fig:separation}
\end{figure}

\section{ Efficient ways to implement bilateral filtering}

Normal conventional bilateral filter as proposed by C.Tomasi and R.Manduchi \cite{tomasi} is very computationally intensive especially when spatial kernel is large. Referring to equation 16 for each pixel $p$ and $q$, whose loop is nested within $p$. So for a 10 megapixel photo there is $100*10^{12}$ iterations involved which make the system very slow and would approximately take more than 10 minutes to process per image.\\
To achieve the filtering there are several efficient approaches which reply on mathematical approximations yielding similar results with acceleration and accuracy that needs to be discussed in the following subsequent sections.

\begin{equation}
g_{\sigma_{ s}}(||q-p ||)g_{\sigma_{ r}}(f(q)-f(p)).(f(q))
\end{equation}

\subsection{Brute Force Approach}

The Brute Force approach follows the conventional way (direct implementation of the bilateral filter) as discussed with equations 16 to 18. Its computational complexity is $O(\left|  S\right|^{2})$  where $\left|S\right|$ the size of the spatial domain (cardinality of $S$) denoting the number of pixels. It's clear that the quadratic complexity enhances the computational cost especially for large spatial domains. So there is a slight modification made with a view to reduce the computational complexity. The idea is to consider only the neighborhood pixels of the central pixel of interest $p$ referring to equation 16 such that $\left||p-q\right||\leq2\sigma_s$ considering the contribution of pixels outside the range of 2$\sigma_s$ is negligible because of spatial kernel. The complexity has been decreased to $O(\left|S\right|\sigma_s^{2})$ and is only effective for small kernels due to its dependency on $\sigma_s$.       
But for this approach the computational complexity increases to a great extent. For a 8 megapixel image   the number of iterations required is $64*10^{12}$ which is obviously not a feasible solution (for the classical approach) for implementation in hardware as the computational complexity increases for processing nested loops of pixels.

\begin{figure}
\centering

\includegraphics[height=5 cm, width=10 cm]{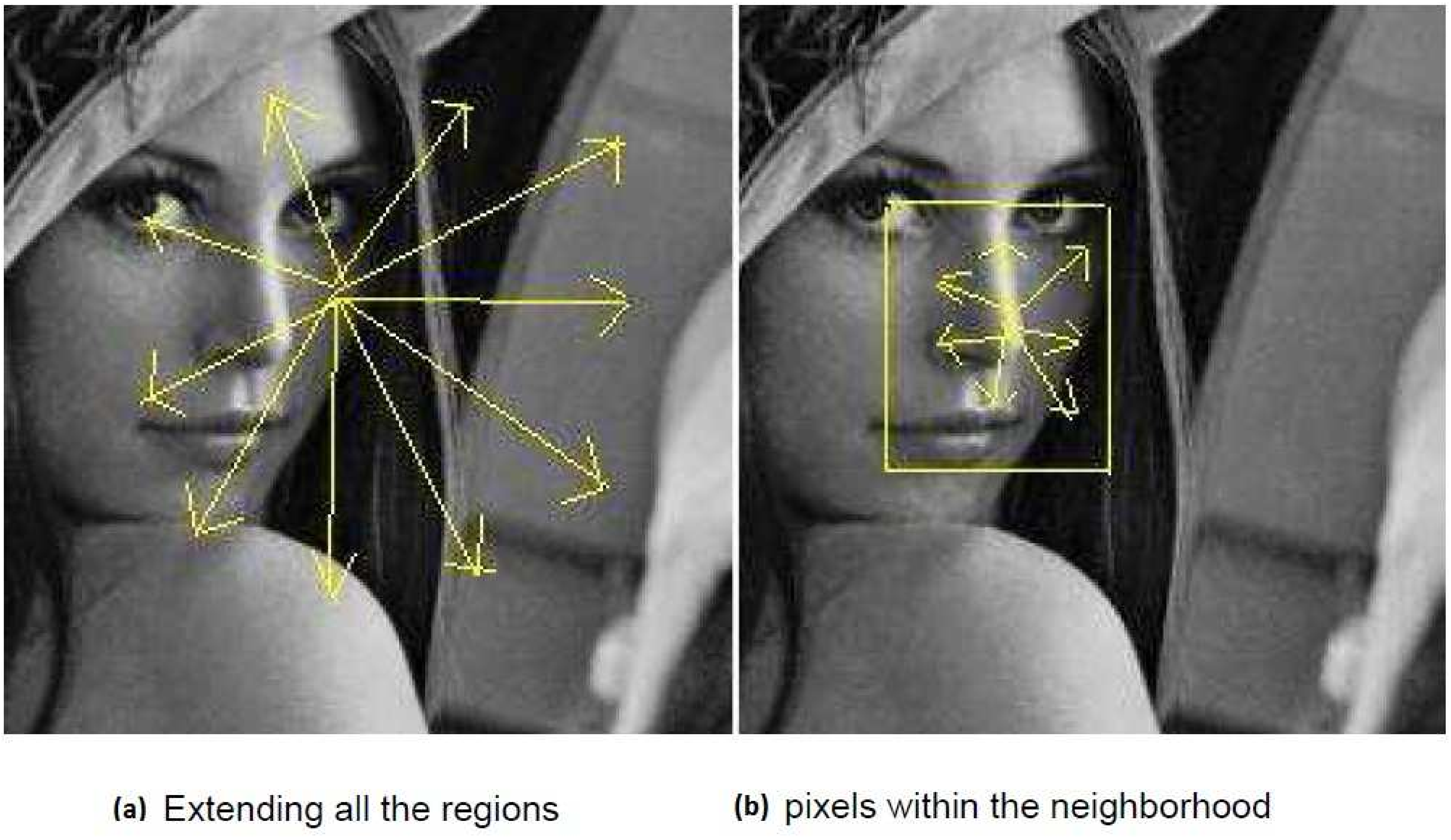}
\caption{The bilateral filters. Each pixel in (a) is replaced by a weighted average of its neighbors as shown in (b). Each neighbor is weighted by a spatial component that penalizes distant pixels and range component that penalizes pixels with a different intensity. Their combination ensures that only nearby similar pixels contribute to the final output. 
Courtesy of \cite{dynamic}.}
\label{fig:iteration}
\end{figure}

\subsection{Jacobi algorithm approach}

In this section we need to discuss the ways to speed-up the bilateral filter and increase it smoothing effect together with its implementation procedure for piece-wise linear signals.
Author \cite{elad} here shown that the bilateral filter can be achieved from the Bayesian approach using a penalty function, and from it a single iteration of the Jacobi algorithm (also known as the diagonal normalized steepest descent DNSD) yields the bilateral filter.
The corresponding penalty function is defined as
\begin{equation}
\varepsilon \left\{ X \right\} = 1/2[X-Y]^T[X-Y]+\lambda/2\sum_{n=1}^{N}[X-D^nX]^TW(Y,n)[X-D^nX].
\end{equation}

where $X$ is the unknown signal and the result should be as close as possible to the measured signal $Y$. The matrix $D$
stands for a one-sample shift to the right operation \cite{elad}. $W$ is the normalized weight matrix. The bilateral filter can be accelerated in the following way:
 
 A quadratic penalty function say:
  \begin{equation}
  \varepsilon\left\{ X \right\}=\sum_{j=1}^{J}[1/2 X^TQ_jX-P^T_jX+C_j]
  \end{equation}
  
  the SD (steepest descent) reads
  
  \begin{equation}
  \hat{X_1}=\hat{X_0}-\frac{\partial \varepsilon\left\{  X\right\}}{\partial X}\mid_{\hat{X_0}}=\hat{X_0}-\mu\sum_{j=1}^{J}[Q_j\hat{X_0}-P_j]
  \end{equation}

The author have shown a close resemblance of equation 19 with 20 by choosing $\left\{ Q_j,P_j \right\}_j$ from 20 as $Q_1= I, Q_k=(I-D^{k-1})^TW(Y,k-1)(I-D^{k-1})$ for $k=2,3,.....,N+1$ AND $P_1=Y, P_1= P_2= P_3= ....= P_{N+1}=0$ for $k=2,3,.....,N+1$.

One effective way to accelerate the SD convergence is the Gauss-Siedel(GS) approach where the samples of $\hat{X_1}$ are sequentially computed from $\hat{X_1}[1]$ to $\hat{X_1}[L]$ ($L$ scalar samples in vector $\hat{X_1}$) and for the calculation of $\hat{X_1}[k]$, updated values of $\hat{X_1}$ are used instead of $\hat{X_0}$ values. This technique is known to be stable and converge to the same global minimum point of the penalty function given in 20. The GS intuition is to compute the output sample by sample and to utilize the already computed values whenever possible. \\
While some describe this process in a more systematic way by decomposition of the Hessian to the upper-triangle, lower-triangle and diagonal parts.  \\
Alternatively the filter can be accelerated by slicing the gradient into several parts. From equation 21 it can be written for $J$ iterations in the following form :

\begin{eqnarray}
\hat{X_1}= \hat{X_0}-\mu[Q_1\hat{X_0}-P_1]  \nonumber \\
\hat{X_2}= \hat{X_1}-\mu[Q_2\hat{X_1}-P_2]  \nonumber \\ 
\hat{X_3}= \hat{X_2}-\mu[Q_3\hat{X_2}-P_3]  \nonumber \\
\cdots \nonumber \\
\cdots \nonumber \\
\cdots \nonumber \\
\hat{X_J}= \hat{X_{J-1}}-\mu[Q_J\hat{X_{J-1}}-P_J]
\end{eqnarray}

Proceeding in this way the final solution $\hat{X_J}$ tends to be closer to the global minimum point of the penalty function in 20. Finally a good de-noising has been achieved with reduced computational complexity but the computational analysis of the complexity has not been shown in the paper\cite{elad}. Well as of its implementation in hardware the iterative steps can be thought of to design as recursive procedure by loop unrolling in hardware.\\

\subsection{Layered Approach to approximate the filter}

Durand et. al \cite{dynamic,parissignal} in order to accelerate the process of filtering, down-sampled the image $I$ at lower resolution followed by computing the product with the range weight $G_{\sigma_r}$ and $G_{\sigma_s}$ the spatial kernel. The final layers are obtained by upsampling the convolution outputs followed by linearly interpolating between the two nearest layers as shown in step 3 of the pseudocode of the Algorithm \ref{algo_bilateral_layered}  which dramatically accelerates the computation.\\
As per Algorithm \ref{algo_bilateral_layered} the downsampling process can be done in software domain itself. However as in step 1 the convolution operation of each layer  $\tilde{L_k}$ with the spatial kernel can be accomplished in parallel with the specialized customized hardware filters meant for convolution operation as discussed in section 3.5 and 3.8 respectively which can accelerate the execution as the layers are independent to each other. Followed by the normalization operations where the pixels need to be divided, CORDIC dividor can be applied thereafter which has got its inherent parallelism for the desired accelerated operation.

\begin{algorithm}[!htb]
\caption{Layered Algorithm as proposed by Durand et. al \cite{dynamic}}
\label{algo_bilateral_layered}
\begin{algorithmic}
     \State  \textbf{Input}: A $2D$ Image I. A set of intensities $\left\{ i_0,i_1,...i_n \right\}$ is chosen and a set of layer $\left\{ \tilde{L_0} ,\tilde{L_1},.....\tilde{L_n} \ \right\}$ is computed as shown:
     $ \tilde L_k=G_{\sigma_r}(\left| i_k-\tilde I_q \right|)\tilde I_q$ where $\tilde{I}$ is a low resolution version of an image computed after downsampling.
     \State 1. Each layer $\tilde{L_k}$ needs to be convolved with the spatial kernel followed by normalisation.\\
   $\tilde L_k=(G_{\sigma_s} \otimes\tilde L_k)\div(G_{\sigma_s}\otimes G_{\sigma_r})$   where the denominator corresponds to the sum of the weights at each pixel and division is the per pixel division. 
     
     \State 2. The layers $\tilde{L_k}$ is upsampled to get $\hat{L_k}.$
     \State 3. For each pixel $p$ with intensity $I_p$, two closest intensity values $i_{k1}$ and $i_{k2}$ needs to be found and the output linear interpolation :\\
     
     $BF[I_p]\approx \frac{I_p-i_{k1}}{i_{k2}-i_{k1}}\hat{L_{k2}}+ \frac{i_{k2}-I_p}{i_{k2}-i_{k1}}\hat{L_{k1}}.$
     
     \State \textbf{Return}: Filtered image $BF[I_p]$.
\end{algorithmic}
\end{algorithm} 

\subsection{Bilateral Grid (3D kernel) approach}

\begin{figure}
\centering
\includegraphics[height=6.2 cm, width=14.5 cm]{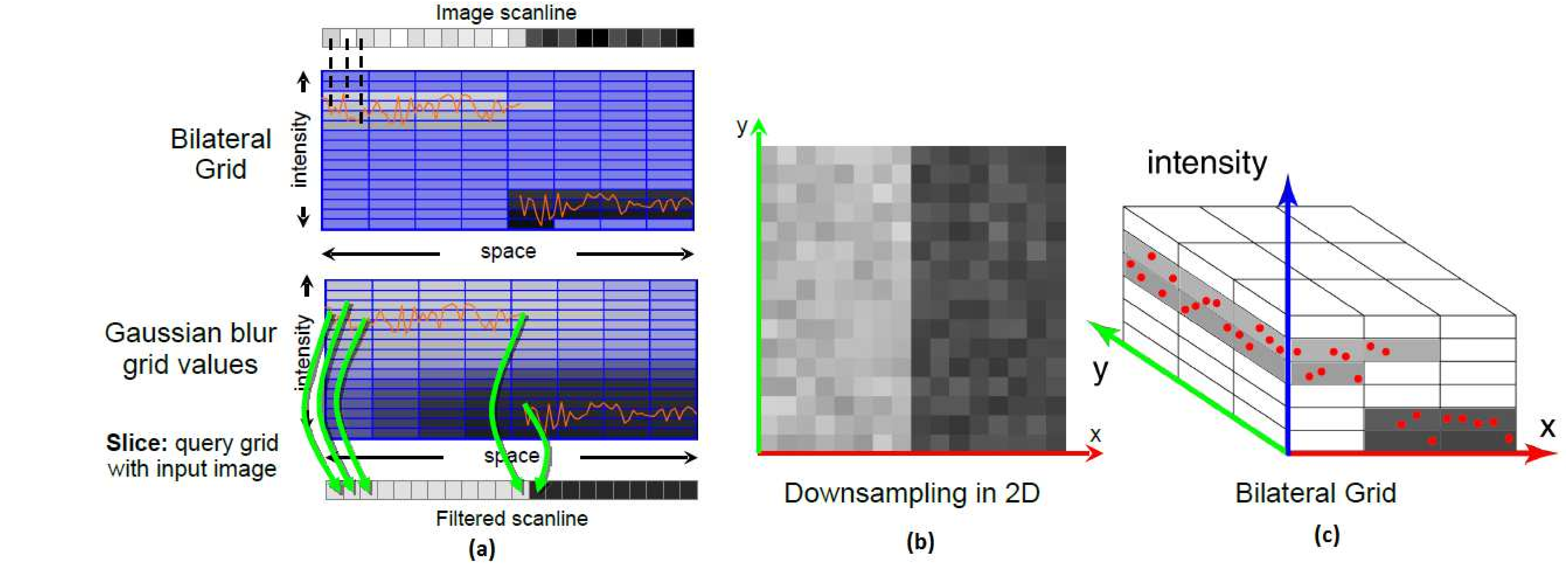}
\caption{The bilateral filters. Each pixel is replaced by a weighted average of its neighbors. Each neighbor is weighted by a spatial component that penalizes distant pixels and range component that penalizes pixels with a different intensity. Their combination ensures that only nearby similar pixels contribute to the final output. Figure reproduced from J.Chen \cite{chen}.}
\label{fig:fig_13}
\end{figure}

\begin{figure}
\centering
\includegraphics[height=12 cm, width=13 cm]{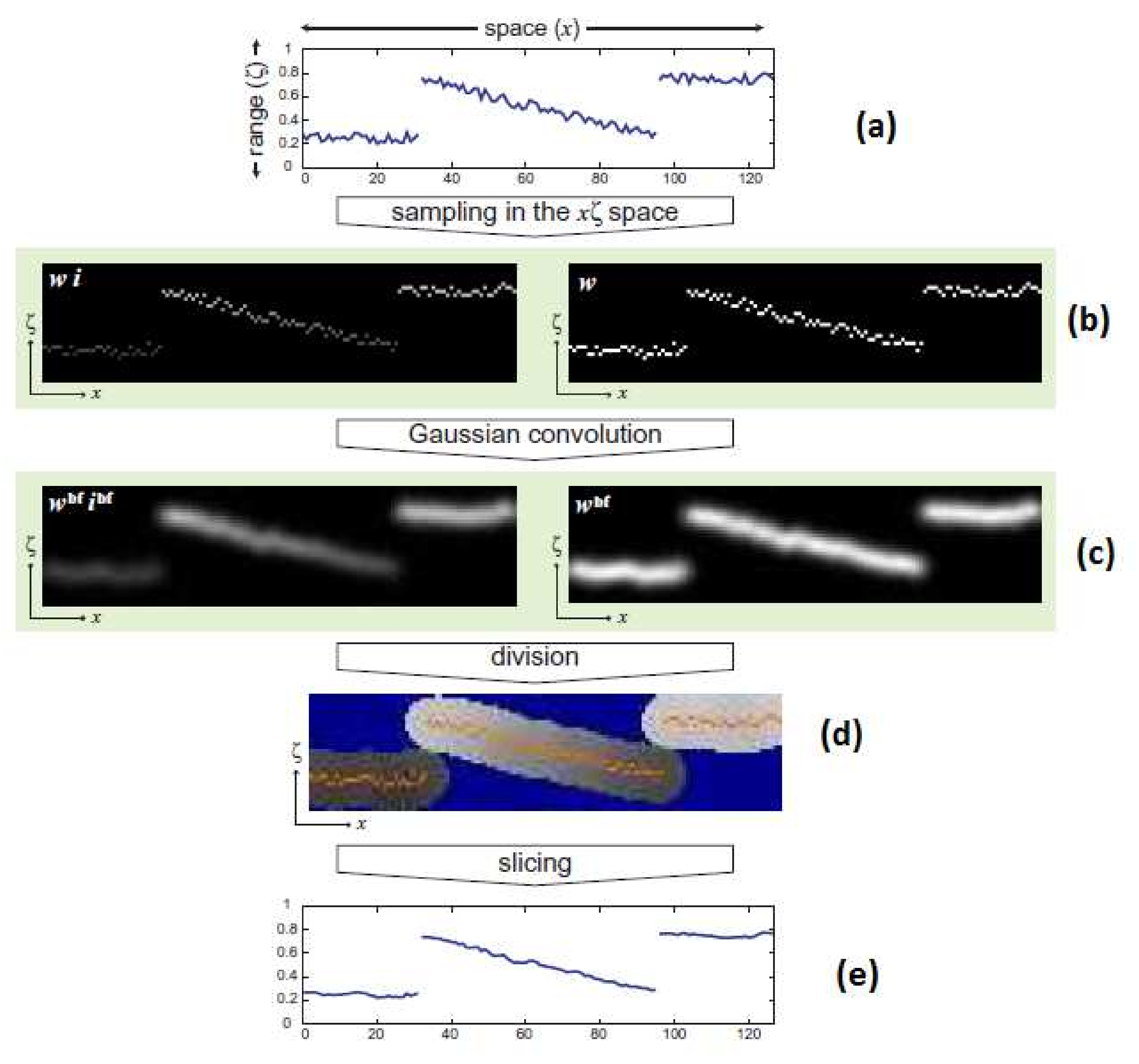}
\caption{The bilateral filters. Each pixel is replaced by a weighted average of its neighbors. Each neighbor is weighted by a spatial component that penalizes distant pixels and range component that penalizes pixels with a different intensity. Their combination ensures that only nearby similar pixels contribute to the final output. Figure reproduced from J.Chen \cite{chen}.}
\label{fig:fig_14}
\end{figure}

J.Chen et. al \cite{chen} evolved with a new idea of data structure to represent image data such that the kernel coefficient weights depends only on the distance between points. The new data structure is the bilateral grid (a volumetric data structure), enabling fast edge preserving image processing. 
Here a 2D image I is represented as a 3D grid $\Gamma$ where the first two dimensions of the grid corresponds to the pixel position $p$ and the third dimension the gray level pixel intensity $I_p$. The total process has been illustrated in Fig.\ref{fig:fig_13} and \ref{fig:fig_14} with the corresponding algorithm in Algorithm   \ref{algo_bilateral_grid}. The complexity of the algorithm is $O(\left|  S\right|+ \frac{\left|  S\right|}{\sigma_s^2}\frac{\left|  R\right|}{\sigma_r})$ where $\left|  S\right|$ and $\left| R\right|$ are the size of the spatial and range domain respectively. There is a problem for color images as it becomes 5D grid and requires large amount of memory.\\ Chen et, al \cite{chen} have already implemented it in real time video rate using  a dedicated graphics processor. Although it computationally very fast still we do not intent to use this procedure of filtering as it does not yield good denoised image quality in signal to noise ratio.

\begin{algorithm}[!htb]
\caption{Bilateral Grid algorithm}
\label{algo_bilateral_grid}
\begin{algorithmic}
     \State  \textbf{Input}: A $2D$ Image I. A grid $\Gamma$ has been build such that $\Gamma:SXR\rightarrow R^2$ containing homogeneous values shown in Fig \ref{fig:fig_13}.a.\\
     $\Gamma(p_x,p_y,r)=(I(p_x,p_y),1)  \  if\ r=I(p_x,p_y)$ \\
    \hspace{1.5 cm} =(0,0) otherwise.
     \State 1. Each layer $\tilde{L_k}$ needs to be convolved with the spatial kernel followed by normalization.\\
   $\tilde L_k=(G_{\sigma_s} \otimes\tilde L_k)\div(G_{\sigma_s}\otimes G_{\sigma_r})$   where the denominator corresponds to the sum of the weights at each pixel and division is the per pixel division shown in fig \ref{fig:fig_14}.c,d. 
     
     \State 2. Downsample $\Gamma$ to get $\tilde{\Gamma}.$ in fig \ref{fig:fig_13}.b and \ref{fig:fig_14}.b respectively.
     \State 3. Perform a Gaussian convolution of $\tilde{\Gamma}$ for each component as shown:\\
     $GC[\tilde{\Gamma}](p_x,p_y,r)=G_{\sigma_s,\sigma_r}\otimes\tilde{\Gamma}(p_x,p_y,r),$
     
     where $G_{\sigma_s,\sigma_r}$ is a 3D Gaussian, $\sigma_s$ and $\sigma_r$ are parameters along the spatial and range dimensions respectively.
     
     \State 4. Upsample $GC[\tilde{\Gamma}]$ to get $\hat{\Gamma}.$

     \State \textbf{Return Result}: For a pixel $p$ with initial intensity $I_p$, $(\tilde{wI},\tilde{w})$ the value at position $(p_x,p_y,I_p)$ in $\hat{\Gamma}$ is denoted. The result of the bilateral filter is 
     $BF[I]_p\approx\tilde{wI}/\tilde{w}. $ 
\end{algorithmic}
\end{algorithm}

\subsection{Separable Filter Kernel Approach}

Pham et al.\cite{pham} presented a separable implementation of the bilateral  filter. The crux of the idea was that multidimensional kernel implementation is computationally expensive. So in order to approximate the 2D bilateral filter it is decomposed into two 1D filters, filtering row followed by column or vice versa. Of course not all kernels are separable.
Say at first the row filtering is applied and to the intermediate results the column filtering is applied to get the same 2D filtered output. The computational complexity decreases linearly from quadratic nature of brute force approach and leads to $O(\left|  S\right|\sigma_s)$ as it neighborhood computation is single dimensional, where $\left|S\right|$ the size of the spatial domain (cardinality of $S$) denoting the number of pixels and $\sigma_s$ defines the neighborhood size of the spatial domain. Separable kernel computes the axis-aligned separable approximation of the bilateral filter which is suitable for homogeneous areas and symmetrical edges but performs poorly in regions of complex textures. There are methods to overcome such problems but it comes with an additional computation overhead. We have successfully implemented the separable kernel based concept of \cite{pham} applied on the trigonometric based kernel \cite{kunalda} on FPGA and obtained a good acceleration as shown in \cite{indicon}.

\subsection{Local Histogram Approach}

Bein Weiss \cite{weiss} computed the bilateral filter from the histogram of image. Considering the spatial weight as a square box function the bilateral filter can be represented as containing only the range kernel component with the domain kernel as a constant entity. Their idea is as follows:
Considering two adjacent pixels $p1$ and $p2$ and their corresponding neighborhoods $N_{\sigma_s}(p1)$ and $N_{\sigma_s}(p2)$. Based on these scenario the histogram of the neighborhood $N_{\sigma_s}(p1)$ is computed keeping an eye with the similarity of the histogram of the other neighborhood.
Once it is known the bilateral filter can be computed as shown:

\begin{eqnarray}
BF[I]_{p1} = 1/W_{p1}\sum_{p2\epsilon N_{\sigma_s}(p1)}^{}G_{\sigma_r}(\left| I_{p1}-I_{p2} \right|)I_{p2} \\
W_{p1}=\sum_{p2\epsilon N_{\sigma_s}(p1)}^{}G_{\sigma_r}(\left| I_{p1}-I_{p2} \right|)
\end{eqnarray}

This implementation if followed by iteration the filter thrice to remove the strange band artifacts near edges of strong intensity gradients which remains after filtering. The complexity of the algorithm is of the order of $O(\left|  S\right|log\sigma_s)$. It can handle filter kernels of any size in short time but has a setback w.r.t processing color images independently for each separate channel. It can exploit vector instructions of CPU and as a result parallel operations of the independent instruction can be implemented concurrently with full acceleration. Interested readers need to refer to the paper \cite{weiss} for the detail explanation of the algorithm. 

\subsection{Constant time bilateral filter implementation using polynomial range kernels.}

Later on Porikli \cite{porikli} in 2008 showed that Gaussian range and arbitrary spatial bilateral filters can be realized by Taylor series as linear filter decomposition with a noticeably good filter response. Considering arbitrary spatial filters a polynomial range has been defined as:

\begin{equation}
g(p+k)=[1-(I(p)-I(p+k))^2]^n
\end{equation}

where $n$ is the order of the polynomial.
Considering bilateral filter of the form 

\begin{equation}
y_b(p)= {k_b}^-1\sum_{k\epsilon S}^{}f(k)I(p+k)g(I(p)-I(p+k))
\end{equation} 

where $g(p,k)$ and $f(k)$ denotes the range and domain kernels respectively and the normalising term $k_b=\sum f(k)g(I(p)-I(p+k)).$ Now for $n=1$, in Eqn  25 and substituting the value in Egn 26 the polynomial range arbitrary spatial filter is obtained as 

\begin{equation}
y_b(p)={k_b}^{-1}[\sum f(k)I(p+k)-I^2(p)\sum f(k)I(p+k)+2I(p)\sum f(k)I^2(p+k)-\sum f(k)I^3(p+k)]
\end{equation}
 
The corresponding linear filter responses are $y_1=\sum f(k)I(p+k)$, $y_2=\sum f(k)I^2(p+k)$, etc. Thereby substituting the above values in Eqn. 27 it can be rewritten as 

\begin{equation}
y_b={k_b}^{-1}[(1-I^2)y_1+2Iy_2-y_3]
\end{equation}  

where $I^1=I(p)$, $I^2=I(p)I(p),$ etc.
Accordingly the normalization term is $k_b=1-I^2+2Iy_1-y_2$ and the spatial filter $f$ is not constrained here resulting for any desired filter to get chosen. For $(n=2)$ Eqn. 25 can be written from Eqn. 26 as 

\begin{equation}
y_b={k_b}^{-1}[(1-2I^2+I^4)y_1+4(I-I^3)y_2+(6I^2-2)y_3-4Iy_4+y_5]
\end{equation}
where ${k_b}^{-1}=1-2I^2+I^4+4(I-I^3)y_1+(6I^2-2)y_2-4Iy_3+y_4$

So its clear from equations 28 and 29 that the bilateral filter has been achieved with polynomial range terms in terms of the spatial filter without approximation. The author had presented one more way namely the Gaussian range. Additional smoothness to realize the bilateral filter by exploiting the property of Gaussian filters to be easily differentiable and can be expressed in terms of linear transforms. The range component is given by

\begin{equation}
exp(-\alpha[I(k+p)-I(p)]^2).
\end{equation}

The equation can be rewritten as 

\begin{equation}
exp(-\alpha I^2(p))exp(-\alpha [I^2(p+k)-2I(p)I(p+k)]).
\end{equation}

Thereby applying Taylor expansion to Eqn. 31 the bilateral filter expansion upto second order and third order derivatives can be obtained as shown:

\begin{equation}
y_b\approx {k_b}^{-1}[y_1 + 2\alpha Iy_2+\alpha (2\alpha I^2-1)y_3-2\alpha^2Iy_4+0.5\alpha^2y_5]
\end{equation}

\begin{equation}
y_b\approx {k_b}^{-1}[y_1+2\alpha I y_2+(2\alpha^2I^2-\alpha)y_3-2\alpha^2(I-\frac{2}{3}\alpha I^3)y_4+\alpha^2(0.5-2I^2\alpha)y_5+\alpha^3Iy_6-(\alpha^3/6)y_7]
\end{equation}

Additively the normalizing terms are similar to some extent containing same terms. So the bilateral filter has been realized using the weighted sum of the spatial filtered responses of the powers of the original image. Their complexity does not scale with the shape or size of the filter and are referred to as $O(1)$ algorithm as its complexity.\\ It is not a  good idea to implement this algorithm in hardware as it would increase the complexity for developing the power series as exponential operation execution is not feasible in hardware. 

\subsection{Constant time bilateral filter implementation using trigonometric range kernels.}

Here the authors Kunal et. al \cite{kunalda} have used the class of trigonometric kernels which turned out to be sufficiently rich, allowing for the approximation of the standard Gaussian bilateral filter. This has been done by generalizing the idea of the using polynomial kernels \cite{porikli}. It has been observed that for a fixed number of terms the quality of approximation achieved using trigonometric kernels is much superior than obtained using polynomial kernels. This was done by approximating the Gaussian range kernel of the bilateral filter using raised cosines. They observed that trigonometric functions share a common property of polynomials which allows one to \emph{linearize} the otherwise non-linear bilateral filter. The trigonometric functions share a common property of polynomial which allows one to linearise the otherwise nonlinear bilateral filter. Recalling the original equation of the bilateral filter in Eqn 16, it is seen that the nonlinear property of the bilateral filter mainly arises due to the range kernel component. If the range part is kept constant and is replaced by a unity value then the filter can be implemented in constant-time, irrespective of the shape or size of the filter. And it is only possible if the range component can be written in the form of a cosine term say

\begin{equation}
g_{\sigma_r}(s)= cos(\gamma s) 
\end{equation}

where $(-T \leq s \leq T)$ and $T$ is the dynamic range of a grayscale image and is equal to 255 and $\gamma=\pi/2T$. The authors have also shown that this idea can also be applied to more general trigonometric ideas. Interested readers can view the details. Now in order to penalize large difference in intensity than small ones it is needed that the family of cosines should be symmetric, non-negative and monotonic. These properties are all enjoyed by the family of raised cosines of the form

\begin{equation}
g_{\sigma_r}(s)= [cos(\gamma s)]^N \hspace{1 cm} where \hspace{.3 cm}   (-T \leq s \leq T)
\end{equation}

Now as the value of the raised power $N$ in Eqn. 35 gets large the $g_{\sigma_r}(s)$ quickly gains Gaussian like behavioral curve over half the period $[-\pi,\pi]$ as $N$ increases. This property speeds up the rate of convergence than that of the Taylor polynomials \cite{porikli} used to approximate the Gaussian range kernel by suitably scaling the Gaussian kernel. 
This approach greatly reduces the computational complexity by $O(1)$ and can also be implemented in parallel further accelerating its speed producing artifact-free results. We like to refer to his article \cite{kunalda} for the detail of the algorithm. We have further improved it w.r.t speed of execution by realizing the filter on an FPGA \cite{indicon}. We have done an FPGA based implementation of an edge preserving fast bilateral filter on a hardware software co-design environment of this most recent algorithm preserving the boundaries, spikes and canyons in presence of noise.
Trigonometric terms (like sine and cosine as in equation 34 and 35)  implementation meant for trigonometric range kernel execution for realizing the fast bilateral filter has been done using the customized circuit (with CORDIC processor) for robust angle computation. Customized CORDIC processor has been designed to compute sine and cosine data with a provision to compute large angles more than the inherent $\pi$ and $2*\pi$ to accommodate/map within it. Followed by the four stage parallel pipelined architecture greatly improves the speed of operation. Moreover, our separable kernel implementation of the filtering hardware increases the speed of execution by almost five times than the traditional convolution filtering, while utilizing less hardware resource \cite{indicon}.
Author in \cite{kunalda} have shown that by exploiting the central property of trigonometric function aka \emph {shiftability} which assists in reducing the redundant computations inherent in the filtering operations. It has been shown that using the shiftable kernels certain complex filtering can be reduced to simply that of computing the moving sum of a stack of images where each image in the stack is obtained through an elementary pointwise transform of the input image adding two significant advantages. Firstly fast recursive algorithms can be used for computing the moving sum and, secondly, parallel computation can be used to further speed up the computation \cite{kunalda,indicon}. 
 \par
\begin{figure}
\centering
\includegraphics[height=3.5 cm, width=16 cm]{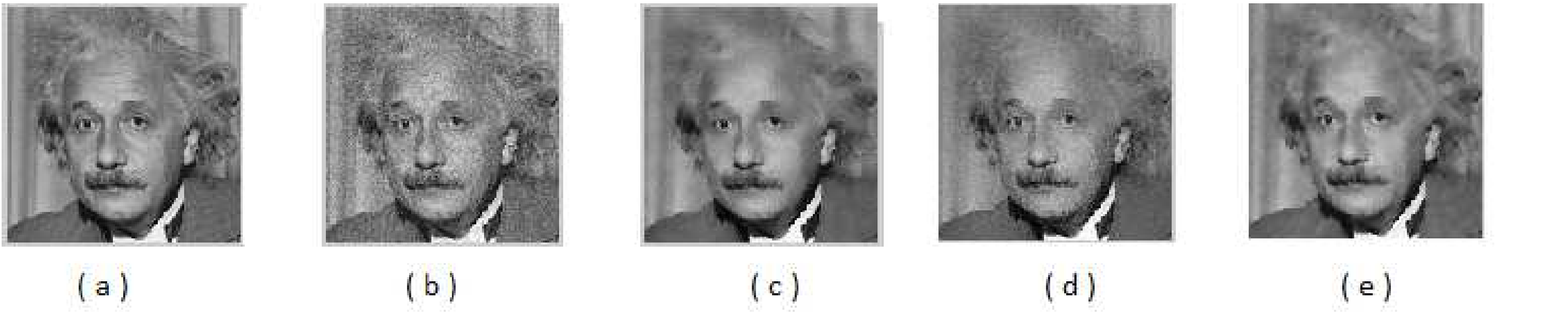}
\caption{(a) denotes the original image, (b) denotes the noisy image with added noise of $\sigma = 0.15, PSNR=20.42 db$, (c) shows the denoised image after applying bilateral filtering with Tomasi$'$s method \cite{tomasi}, $PSNR=21.01 db$, time taken to execute is 1.28 second. (d) shows the denoised image after applying bilateral filtering with Chen's \cite{chen}, $PSNR=20.40 db$, time taken to execute is 0.013 second. (e) shows the denoised image after applying bilateral filtering with Kunal$'$s method \cite{kunalda}, $PSNR=21.14 db$, time taken to execute is 0.75 second.}
\label{fig:fig_15}
\end{figure}

\begin{figure}
\centering
\includegraphics[height=3.8 cm, width=12 cm]{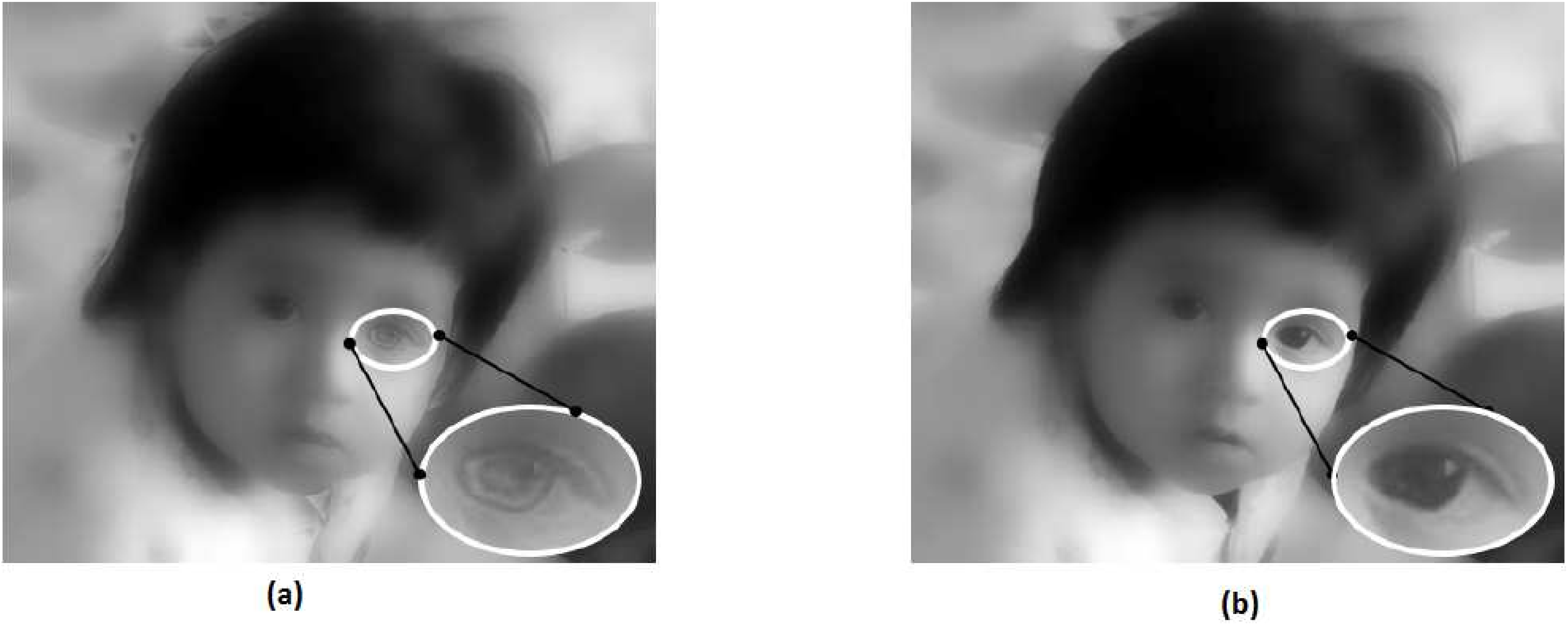}
\caption{The output of two different bilateral filters. The left hand side fig \ref{fig:fig_16}.a shows the output of method \cite{porikli} and right hand side fig \ref{fig:fig_16}.b is of \cite{kunalda}. Note that the strange artifacts in the L.H.S figure particularly around the right eye (see zoomed insets) is due to the approximation caused due to polynomial unlike the one in R.H.S. Figure reproduced from K.N Chaudhury et. al \cite{kunalda}.}
\label{fig:fig_16}
\end{figure}

 Some of the benchmark implementation outcomes has been presented in Figures \ref{fig:fig_15} and \ref{fig:fig_16} respectively. From Fig. \ref{fig:fig_15} we get that the output of Tomasi$'$s method produces a particular $PSNR$ value of the denoised output but consumes a huge time. The outcome of Chen's method in (d) takes the least time of all but with a cost of a decreased $PSNR$ with a slight degraded image quality. (e) shows better results w.r.t both the $PSNR$ and time of execution. 
Also the quality of approximation achieved using trigonometric kernels is much better than obtained using polynomial for a fixed number of terms.
Depending on the availability of the resources, the Bilateral grid approach of Chen \cite{chen} executes well for high resolution real time videos on graphics hardware. The local histogram approach of Weiss \cite{weiss} is better suited for filter kernels of arbitrary size. Finally as discussed above a step ahead to reduce the complexity to $O(1)$ Taylor series polynomial followed by a more advanced version the trigonometric range kernel has been deduced so far. Table \ref{tab:hresult} summarizes the complexity order of each of the previously discussed techniques.

\begin{table}[h]
\caption{Complexity analysis report} 
\centering 
\begin{tabular}{l rrrrrrr} 
\hline\hline 
Algorithm&\multicolumn{1}{l}{Complexity} \\ [0.5ex]
\hline 
Brut Force Approach (section 3.1)  & $O(\left|  S\right|^{2})$\\ 
Layered Approach (section 3.3) & $O(\left|  S\right|+ \frac{\left|  S\right|}{\sigma_s^2}\frac{\left|  R\right|}{\sigma_r})$\\
Bilateral grid(3D kernel) approach (section 3.4) & $O(\left|  S\right|+ \frac{\left|  S\right|}{\sigma_s^2}\frac{\left|  R\right|}{\sigma_r})$\\
Separable filter kernel approach (section 3.5) & $O(\left|  S\right|\sigma_s)$\\
Local Histogram Approach (section 3.6) & $O(\left|  S\right|log\sigma_s)$\\
Constant time polynomial range approach (section 3.7) & $O(1)$\\
Constant time trigonometric range approach (section 3.8) & $O(1)$\\

\hline 
\end{tabular}
\label{tab:hresult}
\end{table}

\section{Applications of edge preserving filters}

This section discusses the the uses of the edge preserving filters (anisotropic and bilateral) for a wide range of applications. 
\subsection{Contrast Enhancement.}

There lies some problems to eliminate the noise from the edges in case of nonlinear isotropic diffusion filter. Anisotropic diffusion do solve the problem as it considers both the modulus of the edge detector $\left|  \bigtriangledown E_{\sigma}\right|$ as well as its direction. 
The authors \cite{stuttgart} constructed an orthonormal system of eigenvectors $v1,v2$ of the diffusion tensor $D$ such that

\begin{equation}
v_{1}||\bigtriangledown E_{\sigma}, \hspace{1 cm} v_2 \perp \bigtriangledown E_{\sigma}
\end{equation}

It is needed to be smoothed along the edge than across it. As a result Weickert \cite{Weickert} choosed two eigenvalues $\lambda_1$ and $\lambda_2$ such that

\begin{eqnarray}
\lambda_1(\bigtriangledown E_{\sigma}):=g(\left| \bigtriangledown E_{\sigma} \right|^2)\\
\lambda_2(\bigtriangledown E_{\sigma}):=1
\end{eqnarray}

The goal of the filter decides the choice of the eigenvalues $\lambda_1$ and $\lambda_2$. 
Smoothing within each region and inhibition of diffusion across edges can be obtained by reducing the diffusivity $\lambda_1$ perpendicular to edges. Increasing it enhances the contrast. It depicts the changes in the Gaussian like function with temporal evolution. This edge location remains stable over some time till the process of contrast enhancement is concluded the steepness of edge decreases and the image converges towards a constant image.

\begin{figure}
\centering
\includegraphics[height=9 cm, width=9 cm]{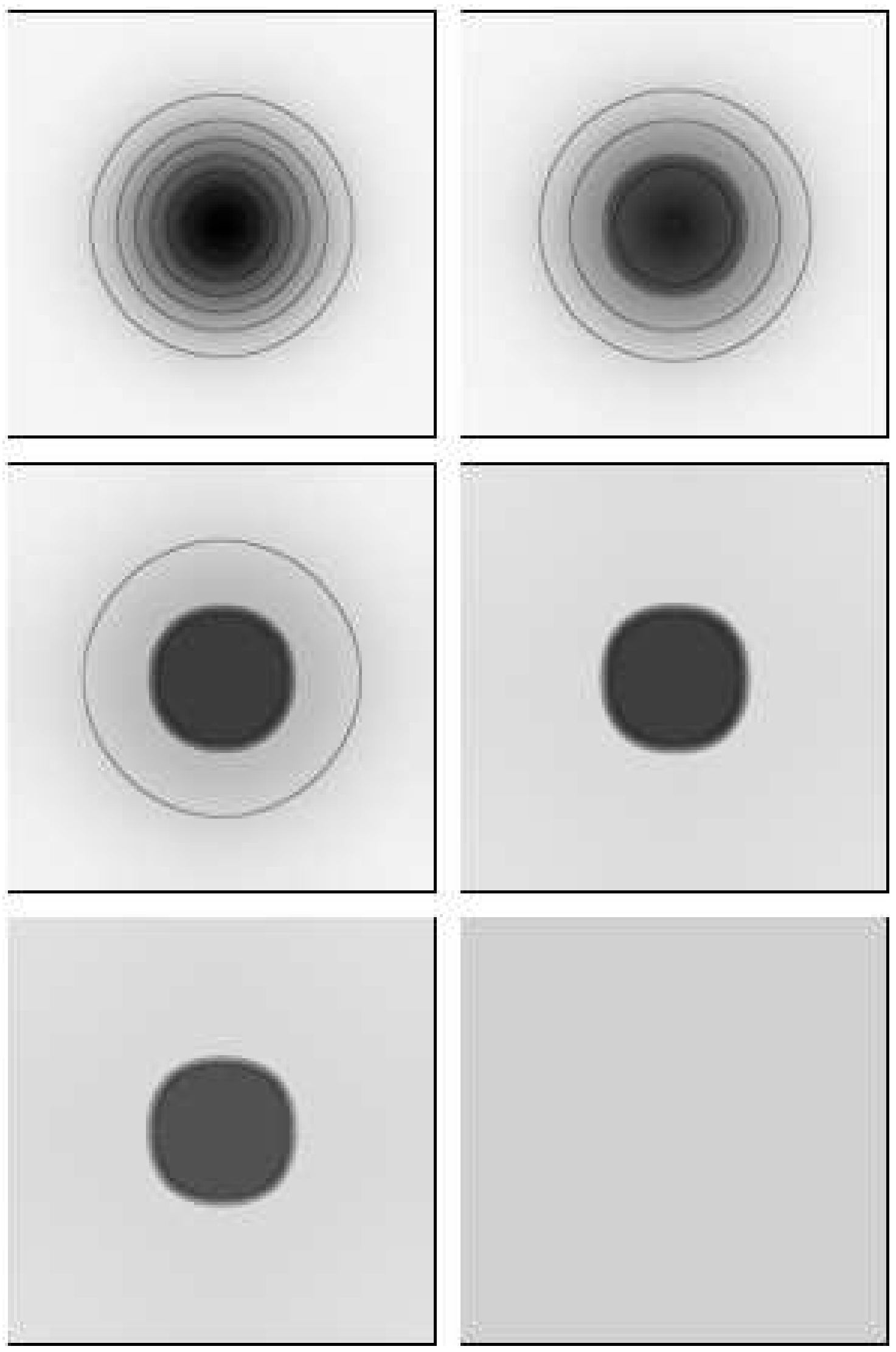}
\caption{Anisotropic diffusion filtering applied on a Gaussian type function with parameters $\sigma = 2, \lambda=3.6$ and $\Omega = (0,256)^2$. Top to bottom all the figures are denoted with time beginning from t=0, 125, 615, 3125, 15625, 78125. Figure reproduced from B.Stuttgart et. al \cite{stuttgart}.}
\label{fig:fig_17}
\end{figure} 

\subsection{Coherence enhancement}

\begin{figure}
\centering
\includegraphics[height=5 cm, width=10 cm]{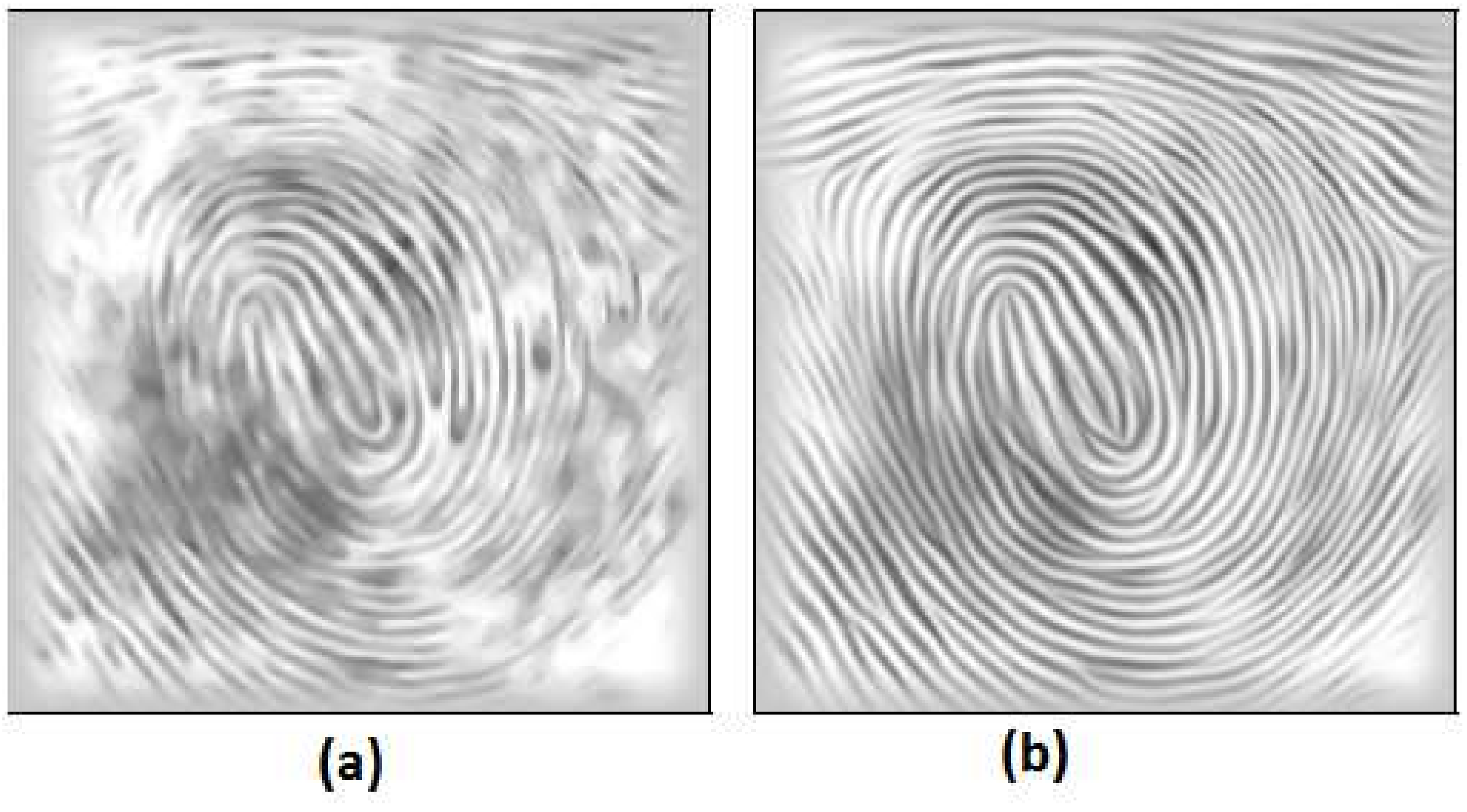}
\caption{Anisotropic diffusion filtering applied on a fingerprint image. (a) image shows the distorted noisy input image while (b) the coherence enhanced image applied with parameters, $\sigma=0.5, \rho=4, t=20$. Figure reproduced from B.Stuttgart et. al \cite{stuttgart}.}
\label{fig:fig_18}
\end{figure}

It begins with the wish to process one-dimensional features like line-like features \cite{stuttgart}. Cottet et. al \cite{cottet} with their diffusion tensor $D$, eigenvectors as in Equation 36, with the eigenvalues as shown:

\begin{eqnarray}
\lambda_1(\bigtriangledown E_{\sigma}) = 0\\
\lambda_2(\bigtriangledown E_{\sigma}) = \frac{\eta\left| \bigtriangledown E_{\sigma} \right|^2}{1+(\left| \bigtriangledown E_{\sigma} \right|/\sigma)^2}   \hspace{1 cm}  (\eta > 0).
\end{eqnarray}

which denotes the diffusion process diffusing perpendicular to $\bigtriangledown u_{\sigma}$. As $\sigma \rightarrow 0$, $\bigtriangledown u$ becomes an eigenvector of $D(J_{\rho}(\bigtriangledown u_{\sigma}))$ with the corresponding eigenvalue diminishing to 0, halting the process completely. Now by substituting $\bigtriangledown E_{\sigma}$ by a more robust local descriptor one can improve the parallel line like textures thereby enhancing the coherence-enhancing anisotropic diffusion. Figure \ref{fig:fig_18} shows the results of the AD filtering applied on a fingerprint image.

\subsection{Denoising}

Both anisotropic as well as bilateral filtering are used for denoising purpose. Here in this section we try to focus more on the use of bilateral filter to remove noise from digital images.\\
There are several sources of noise corrupting a digital image like dark current noise \cite{link} due to thermally generated electrons at the sensor site, shot noise arising out of quantum uncertainty in photo-electron generation. Amplifier noise and quantization noise occur during the conversion of the number of electrons generated to pixel intensities \cite{link}. The basic purpose of the bilateral filter is image denoising and has a wide range of applications like image restoration, medical imaging, object tracking etc. to name a few.
Bilateral filtering is very simple, easy to understand and its recent implementations are very fast. It preserves object contours (edge preserving) and generates sharp results. It smooths the homogeneous intensity regions while respecting the edges and in this way it reduces the noise injected in images but sometimes the edge preservation is not accurate and introduces undesirable \emph{staircase effect} and cartoon-like features. It is undoubtedly a good solution but not for all kinds of application. The application areas where it has been applied are :

\subsubsection{Normal denoising}

\begin{figure}
\centering
\hspace*{1pt}
\includegraphics[height=7 cm, width=11 cm]{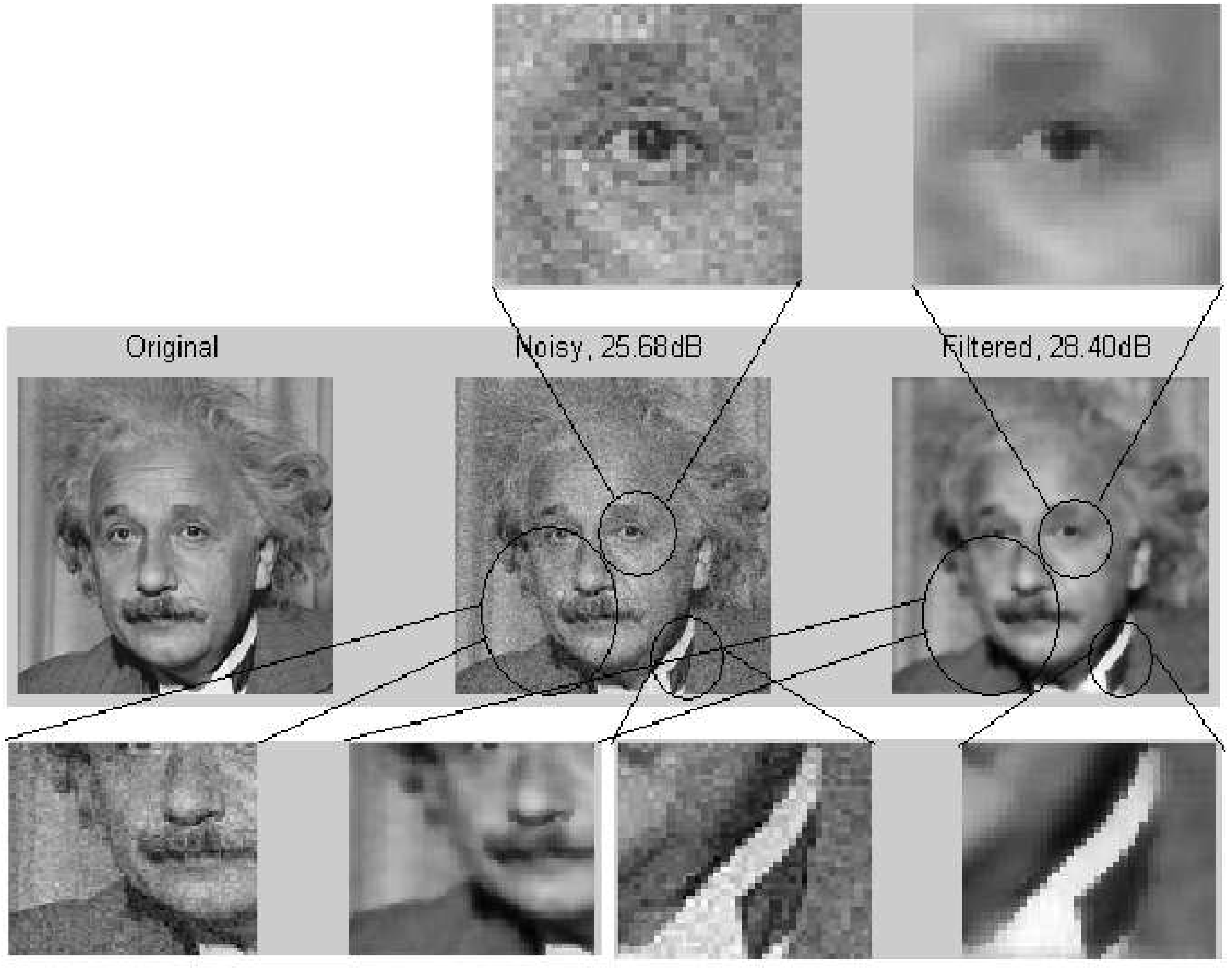}
\vspace*{-8pt}
\caption{Filter output for image size of $150*150$ for the additive Gaussian noise. Filter
settings $\sigma_s$=20, $\sigma_r$=50 and $\sigma$=12 for the additive Gaussian noise. Figure reproduced from \cite{indicon}}
\label{fig:fig19}
\end{figure}

\begin{figure}
\centering
\includegraphics[height=8 cm, width=13 cm]{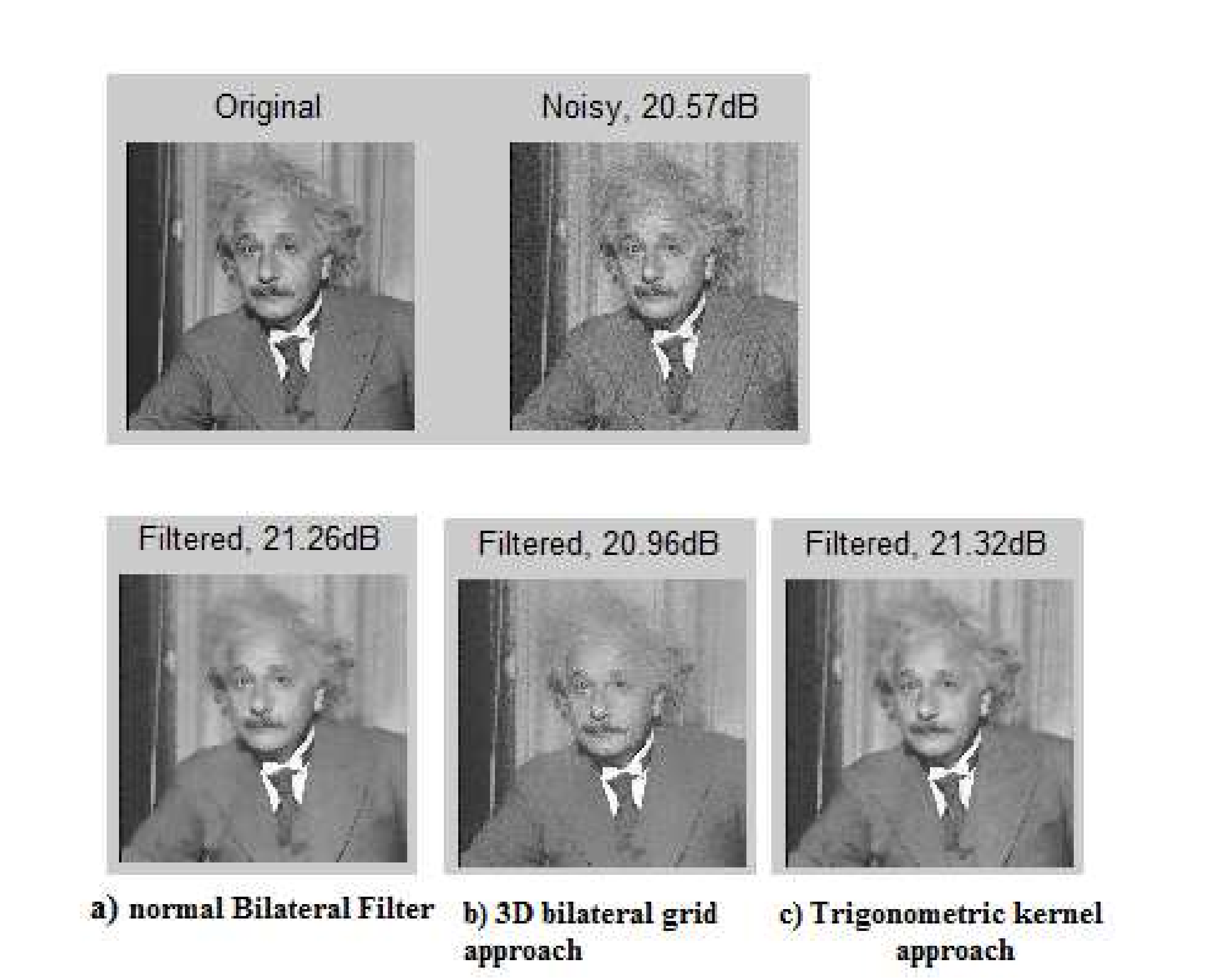}
\caption{PSNR values of the filtered images for the different Gaussian Bilateral filters on the grayscale image of Einstein of size $150*150$ with filter settings $\sigma_s$= 16, $\sigma_r$= 0.1 and $\sigma$= 0.15 for the additive Gaussian noise. Figure reproduced from \cite{indicon}.}
\label{fig:fig20}
\end{figure}

As we know that bilateral filtering is used for denoising purposes. We like to validate it via some experimental results as shown in Figure \ref{fig:fig19} and \ref{fig:fig20} with the parameters correspondingly as shown. Figure \ref{fig:fig19} shows the original image with its corresponding noisy one after adding white Gaussian noise. The right hand side contains its filtered version with zoomed insets surrounding it to provide  a clear vision. Figure \ref{fig:fig20} shows the filtered version applying various benchmark methods starting from normal classical implementation as in $a$, $b$ shows the 3D grid implementation and $c$ shows the trigonometric kernel implementation \cite{kunalda}. The corresponding denoised images are also visible and they are used as per the demand in applications.

The experiment has been implemented 10 times in software and the corresponding mean squared error ($MSE$) obtained has been averaged by 10 to get the averaged $MSE_{av}$, which is used to calculate the $PSNR$ of the software.

\subsubsection{Video implementation}

T.Pham \cite{pham},  J. Chen \cite{chen} Bennet et. al \cite{bennett} shows that bilateral filter can be implemented for real time videos. T. Pham have shown that better image quality and and compression efficiency can be achieved if the original video is preprocessed with the separable filter kernel computing at a fraction of executing time unlike the traditional one. Chen et. al have shown the use of a new data structure the \emph{bilateral grid} approach already discussed in section 3.4. Using it edge aware image manipulations such as local tone mapping on high resolution images can be done at real time frame rate with parallelising the algorithmic instructions on modern GPUs.

\begin{figure}
\centering
\includegraphics[height=5.6 cm, width=12 cm]{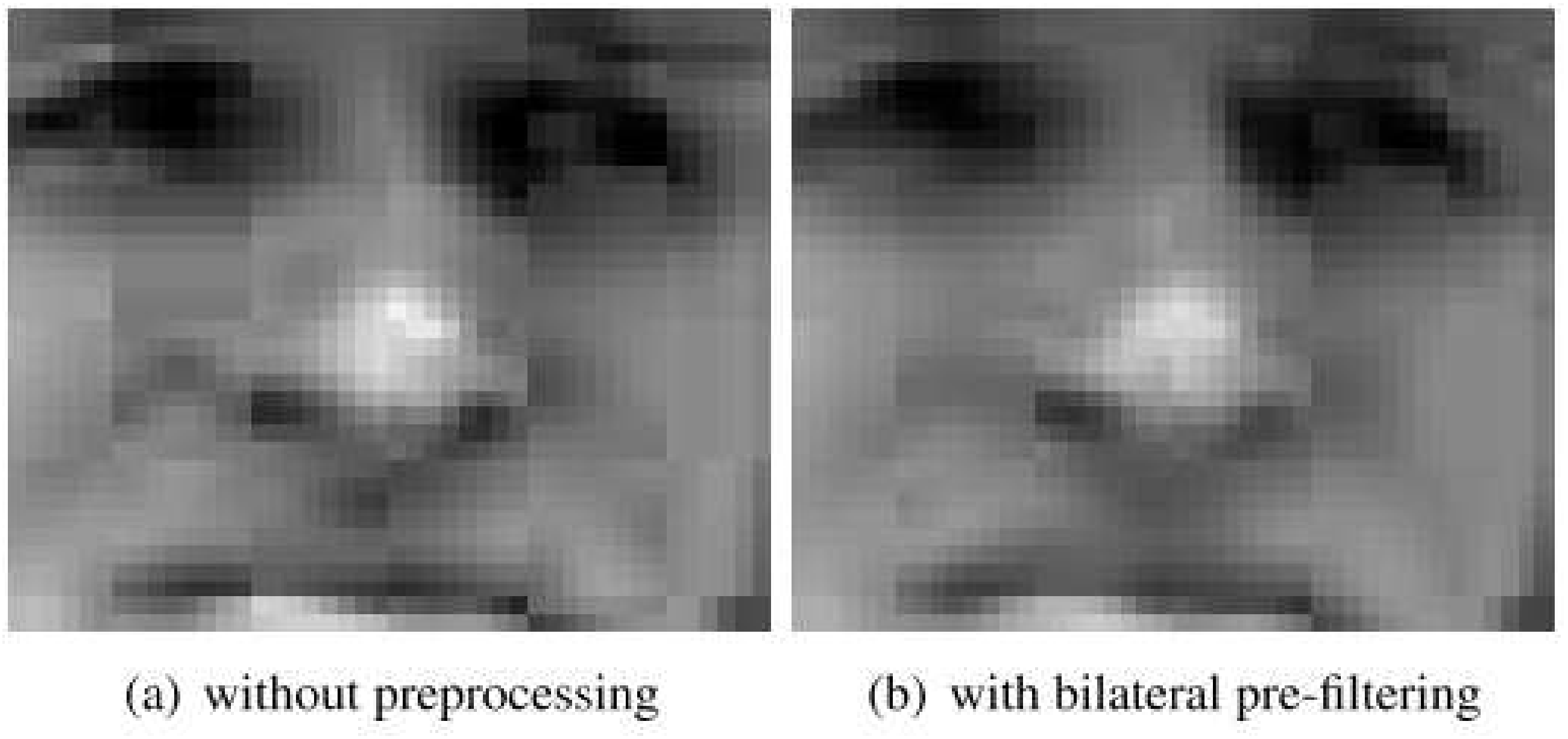}
\caption{Frame 31 of a Foreman sequence compressed at 150 Kbits/s where the video without preprocessing on the left in (a) is more blocky than its counterpart shown in (b). The process is described in section 3.5. Figure reproduced from \cite{pham}.}
\label{fig:fig21}
\end{figure}

\begin{figure}
\centering
\includegraphics[height=4 cm, width=14 cm]{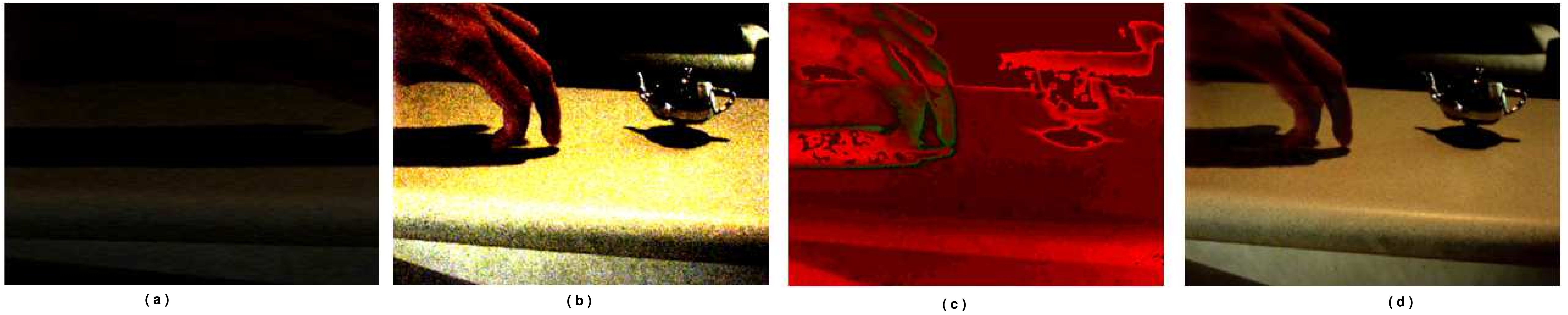}
\caption{(a) Input frame, (b) Histogram stretched version, (c) red and green shows the number of temporal and spatial pixels integration, (d) Output of Bennett et. al \cite{bennett}. It describes the process to combine spatial and temporal bilateral filtering to achieve high-quality video denoising and exposure correction. Figure reproduced from \cite{bennett}.}
\label{fig:fig22}
\end{figure}

\subsubsection{Medical Image Processing}

In medical image processing Wong et. al \cite{wong} and D. Bhonsle et. al \cite{ijigsp} removes the noises corrupting the images mainly in presence of white Gaussian noise. Although it doesn't work good with \emph{salt and pepper} noise. Wong improved the structure preservation abilities of the bilateral filter by an  additional weight component depending on the local shape and orientation known as trilateral filtering.

\subsubsection{Orientation smoothing}

D.Chen et. al \cite{lenkaji} introduced first the concept of \emph{orientation bilateral filter} which combines the concept of average(arithmetic mean value) orientation and the bilateral filter to maintain microstructural features and to
smooth orientation noise which needs a fast computation of the disorientation using quaternion equation by Hamiltonian as follows:

\begin{equation}
Q=cos(\theta/2), \hspace{.5 cm} r_1sin(\theta/2),\hspace{.5 cm} r_2sin(\theta/2),\hspace{.5 cm} r_3sin(\theta/2)
\end{equation}

\begin{figure}
\centering
\includegraphics[height=9 cm, width=14 cm]{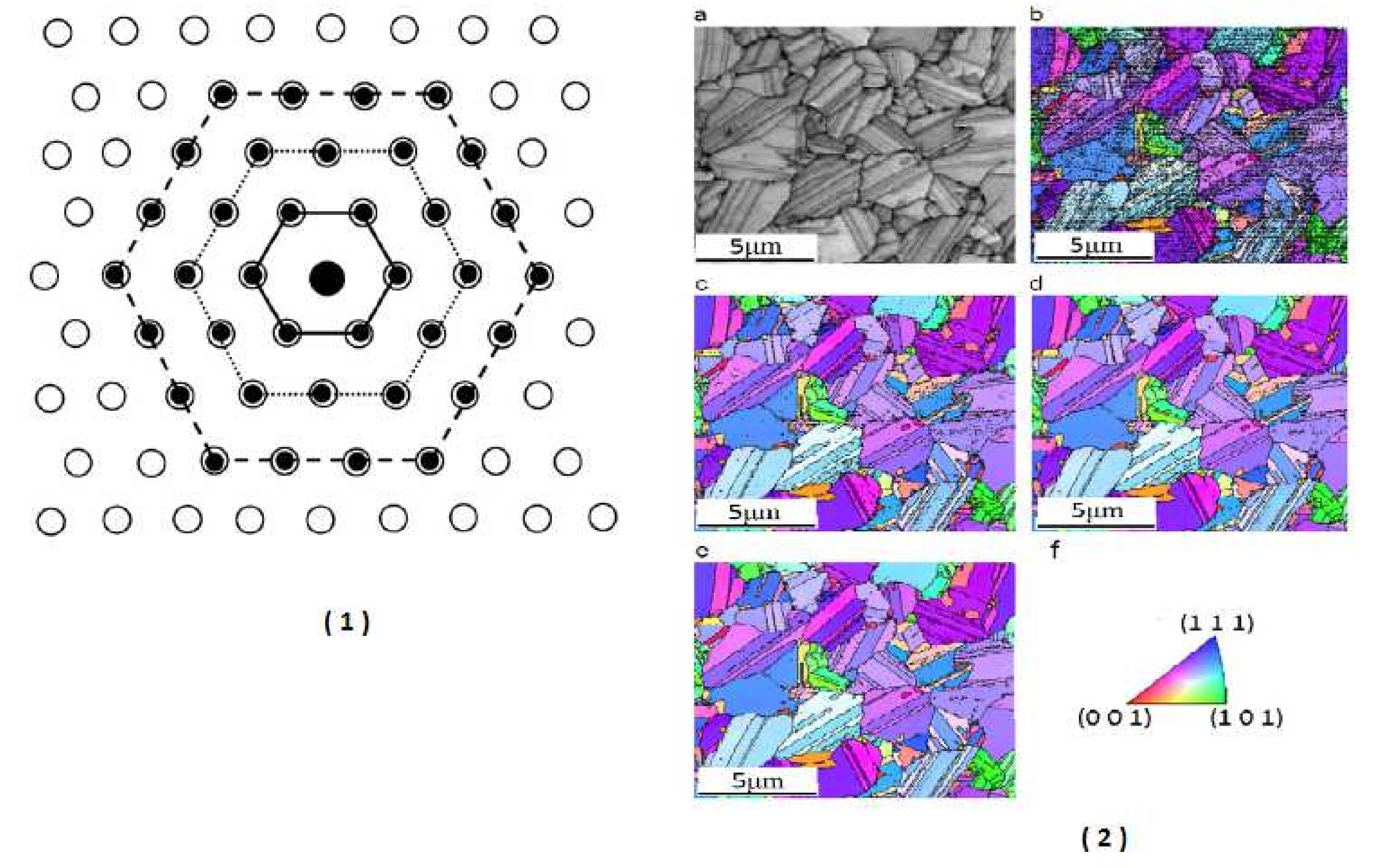}
\caption{(1) Hexagonal sampling grids meant for the bilateral filtering process of a hexagonal image structure, (2) (a) orientation maps for the cross-section of
electro-deposited copper using inverse pole figure color coding with the reference
direction in the normal direction (ND): (b) without a filtering process, (c) using a
hexagonal sampling grid of 7 sampling points, (d) using a hexagonal sampling grid
of 19 sampling points, (e) using a hexagonal sampling grid of 37 sampling points,
and (f) the inverse pole figure color coding. Figure reproduced from \cite{lenkaji}.}
\label{fig:fig23}
\end{figure}
\begin{figure}
\centering
\includegraphics[height=5 cm, width=14 cm]{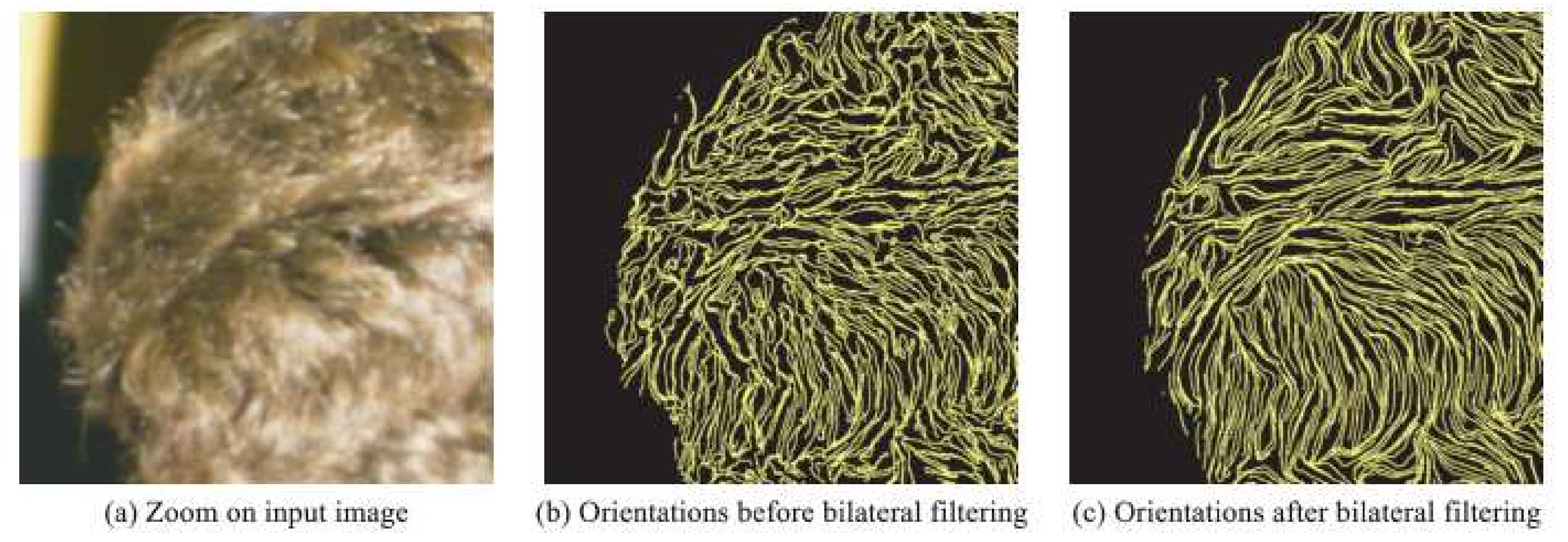}
\caption{Hair orientation measurements with applied bilateral filter to a complex plane. Figure reproduced from Paris et. al \cite{parishair}.}
\label{fig:fig24}
\end{figure}


where $r=(r_1,r_2,r_3)$ denotes the is the axis and $\theta$ is the angle of rotation of the axis which completes the orientation specification. The average orientation based on the orientation filter is given by the weighted function:

\begin{equation}
Q(x_0,y_0)=\frac{\sum_{i=0}^{n-1}Q(x_i,y_i)*W(x_i,y_i)}{\sum_{i=0}^{n-1}W(x_i,y_i)}
\end{equation}

where the weight $W(x_i,y_i)$ is the product of two components namely the domain and range part respectively as given by $W_S(x_i,y_i)$ and $W_R(x_i,y_i)$ which are obtained by methods described in section 2.3.1 of this paper. The intensity difference of the range component here denotes the misorientation between the
sampling point $(x_i,y_i)$ and the center point $(x_0,y_0)$. the smoothing process is continued as per the equation 42 and the output is as shown in Figure \ref{fig:fig23}.
Through this the authors have tried to improve the angular precision of orientation maps for deposited and deformed structures of pure copper obtained from electron backscattered diffraction (EBSD) measurements used in the quantitative characterization of crystallographic microstructures.\\ 
Paris et. al used also applied the same filter to smooth the 2D orientation field from optical measurements for hairstyle modeling \cite{parishair} as shown in Figure \ref{fig:fig24}.

\subsection{Contrast management}

Bilateral filtering can be applied for contrast management applications like grouping large and small scale components of an image (detail enhancement) or reduction, texture and illuminating separation, tone mapping and management and generation of high dynamic range imaging \cite{elad}.  

\subsubsection{Decomposition for detail enhancement and Image fusion}

\begin{figure}
\centering
\includegraphics[height=4 cm, width=13 cm]{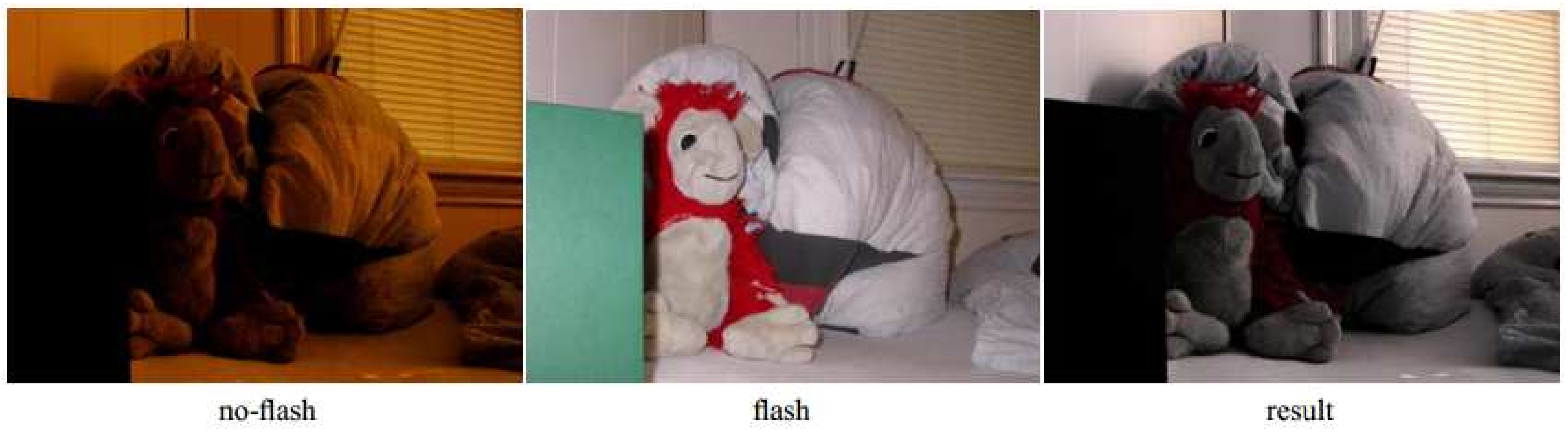}
\caption{Figure showing the no-flash,flash and the resultant output of the previous duo. Figure reproduced from Elmar et. al \cite{elmar}.}
\label{fig:fig25}
\end{figure}

For detail enhancement Fattal et. al \cite{fattal} used a new image-based technique for enhancing the shape and surface details of an object. They computed 
a multiscale decomposition based on the bilateral filter for a set of photographs taken from a fixed viewpoint, but under varying
lighting conditions as an input. For a basic multiscale bilateral decomposition its purpose is to build a series of filtered images $I^j$ that preserve the strongest edges in $I$ while smoothing small changes in intensity. At
the finest scale $j = 0$, $I^0 = I$ is set and then iteratively the
bilateral filter is applied to compute:

\begin{eqnarray}
I_p^{j+1}=\frac{1}{k}\sum_{q\epsilon \Omega }^{}g_{\sigma_{s,j}}(\left\Vert q \right\Vert)g_{\sigma_{r,j}}(I_{p+q}^j-I_p^j)I_{p+q}^j\\
k= \sum_{q\epsilon \Omega }^{}g_{\sigma_{s,j}}(\left\Vert q \right\Vert)g_{\sigma_{r,j}}(I_{p+q}^j-I_p^j)
\end{eqnarray}

where $p$ is the pixel coordinate, $\sigma_{s,j}$ and $\sigma_{r,j}$ are the widths of the spatial and range Gaussians respectively and
$q$ is an offset relative to $p$ that runs across the support of the spatial Gaussian. Finally an enhanced
image combining detail information at each scale across
all the input images needs to be reconstructed. It is simple to implement, fast $O(N^2logN)$ and accurate. Results shown in Figure \ref{fig:fig25}. It is to be noted that the bilateral filter is a good choice for multiscale implementation as it avoids the halo artifacts commonly associated with the classical Laplacian image pyramid. 
\par Image fusion is again categorized into two applications namely (a) flash photography and (b) fusion of multispectral images. \\
\emph{Flash photography}: Visually compelling images can be generated in low lighting conditions by combining flash and no-flash photographs. Elmar et. al \cite{elmar} used the bilateral filter to decompose the image into detail and large scale. The finer textures(detailed feature) extracted from the flash photography is combined with the large scale components of no-flash image. \\
\emph{Multispectral fusion}:  Here both the data from infrared spectrum and standard RGB data is used to denoise video frames in low light conditions as per Bennett et al.\cite{bennett}. A modified bilateral filter (dual bilateral filter) is applied in such condition where the range weight is constituted by a combination of both the visible spectrum and the infrared spectrum given by:

\begin{equation}
BF_d[RGB]_p=1/W_p\sum_{q\epsilon S}^{}G_{\sigma_s}(\left\Vert p-q \right\Vert)G_{\sigma_{RGB}}(\left\Vert RGB_p-RGB_q \right\Vert).G_{\sigma_{IR}}(\left\Vert IR_p-IR_q \right\Vert)RGB_q
\end{equation} 

where $RGB_p$ is a 3-vector representing $RGB$ component at pixel $p$ and $IR_p$ the intensity of the infrared spectrum at the same pixel $p$. 

\begin{equation}
BF^{nf}=\frac{1}{k(s)}\sum_{p\epsilon \Omega}^{}f(p-s)g(I_p^f-I_s^f)I_p^{nf}
\end{equation}

where $f$ and $nf$ denotes the flash and no-flash pictures.
and $f(g-s)$ denotes the domain kernel and $g(I_p^f-I_s^f)$
denotes the range kernel. Here the edge preserving term although corrupted in the no-flash image is preserved.

\subsubsection{Large and small scale component separation}

It has been observed that for large range kernel standard deviation $\sigma_r$, the bilateral filter eliminates the fine texture variations of reflectance while maintaining large discontinuities of illumination changes \cite{oh}. The crux of the idea is illumination changes occur  at a large scale than texture patterns.

\begin{figure}
\centering
\includegraphics[height=5 cm, width=13 cm]{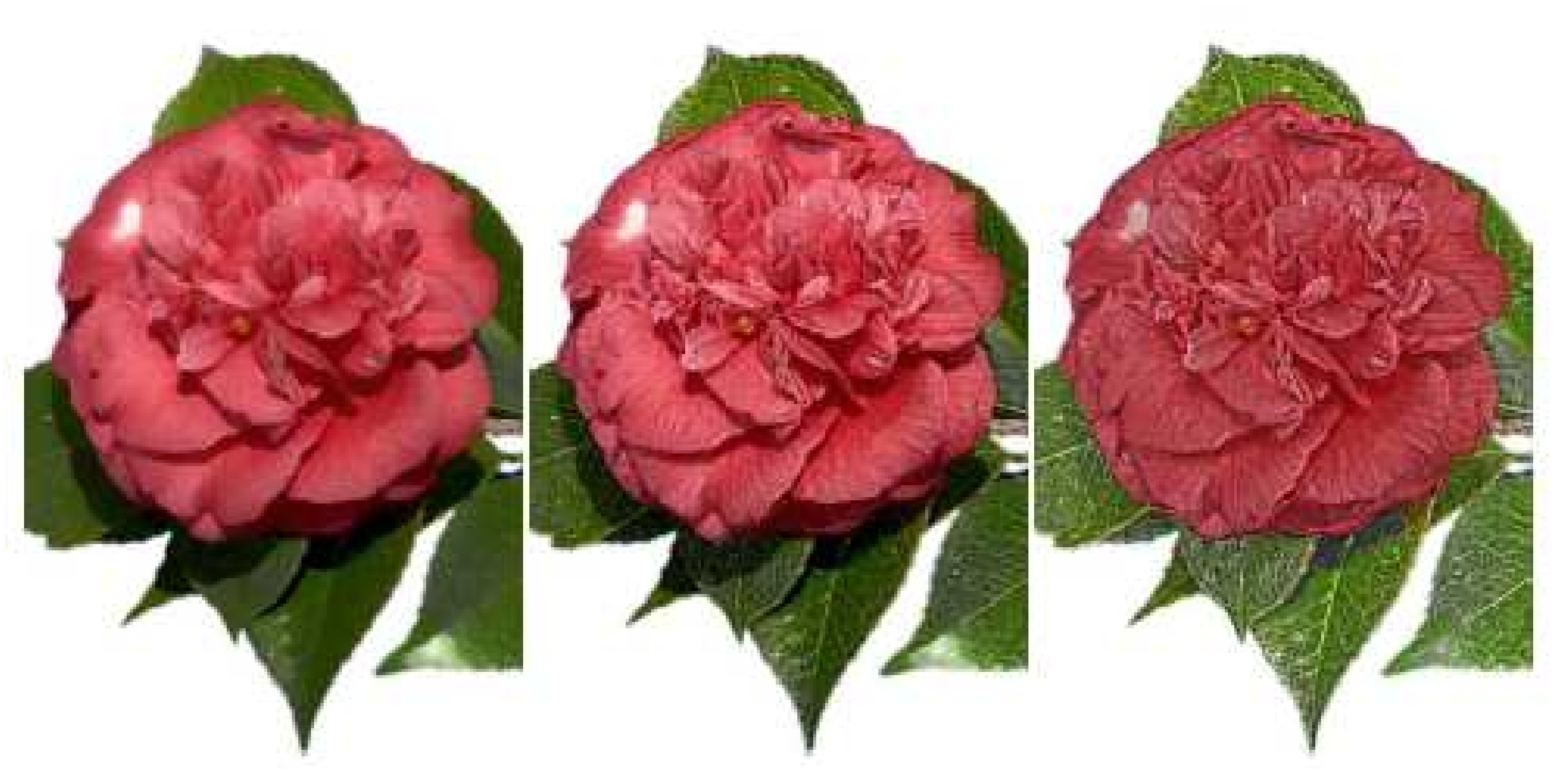}
\caption{Usage of bilateral filter to create multiscale decomposition of images revealing minute details after decomposition and combining shading information across all the images. Figure reproduced from Fattal et. al \cite{fattal}.}
\label{fig:fig26}
\end{figure}
 
\subsubsection{Tone Mapping and management}

Durand et. al \cite{dynamic} in the year 2001 demonstrated the significant property of bilateral filter to seclude small-scale signal variations, can be utilized to design a tone mapping process which can map intensities of high dynamic range image to a low dynamic range display. They overcame the limitations of the existing classical approach like gamma reduction, uniform scaling etc. which wipes out the finer details and texture depths of displays. Here the bilateral filter is applied on the log intensities of the high dynamic range images, followed by evenly scaling down the result and summing up filtered residual, resulting in a more visually pleasing display \cite{dynamic}.
The texture of the image is defined by:

\begin{equation}
\frac{1}{W_p}\sum_{q\epsilon S}^{}G_{\sigma_s }(\left\Vert  p-q\right\Vert)G_{\sigma_r}(\left|log I_p-logI_q  \right|)\left| H \right|_q
\end{equation}  

where the texture can be varied by adjusting the small scale component visible in the image. For contrast management purpose the bilateral filter is usually applied to the log of the source image as the response of human visual sense is multiplicative. Also logarithm operation helps the standard deviation of range component to act uniformly across different levels of intensity. However 
using a cubic root which is based on luminance channel of the CIE-Lab color space, handles more efficiently the multiplicative process due to its computational simplicity.

\begin{figure}
\centering
\includegraphics[height=5 cm, width=15 cm]{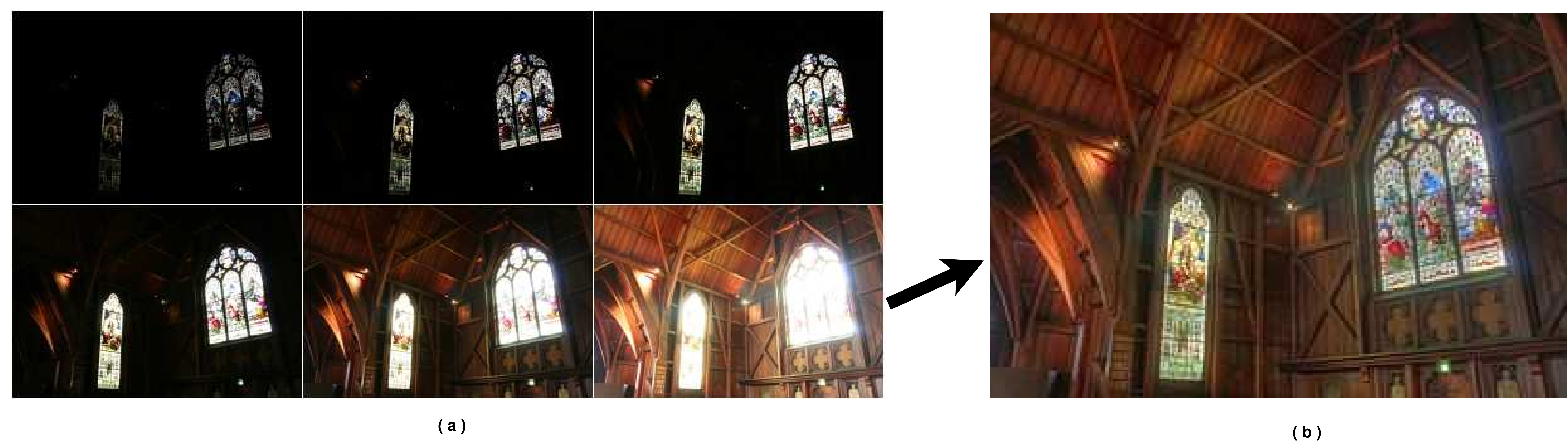}
\caption{Usage of bilateral filter to create a tone mapped image. The six individual images in (a) is used to generate the image as shown in the right hand side (b). It is to be noted that the fine details of interior of the church is not visible due to underexposed and overexposed regions in light, and grabbing all the information from the six images from (a), figure (b) is generated where almost all of the fine details(depth) are visible. Figure reproduced from wiki.\cite{tone}.}
\label{fig:fig27}
\end{figure}


\subsection{Depth Reconstruction}

Bilateral filter improves in depth reconstruction, i.e recovery of pixel depth value from corresponding pixels point pairs of different views. Q.Yang et. al \cite{yang75} constructed a metric called \emph{cost volume}, which is a 3D volume of depth probability, provided by the probabilistic distribution of depth of input range map image. Then they applied iteratively the bilateral filtering on the cost volume to generate visually compelling high resolution range images with accurate depth estimate exhibiting a very effective but simple formulation. Yang \cite{yang76} again in the year 2009 showed the usage of bilateral filter in stereo reconstruction.  
Among several strategies for depth reconstruction bilateral aggregation turned out to be the most successful one.

\subsection{Feature preserving mesh smoothing}

\begin{figure}
\centering
\includegraphics[height=3 cm, width=14 cm]{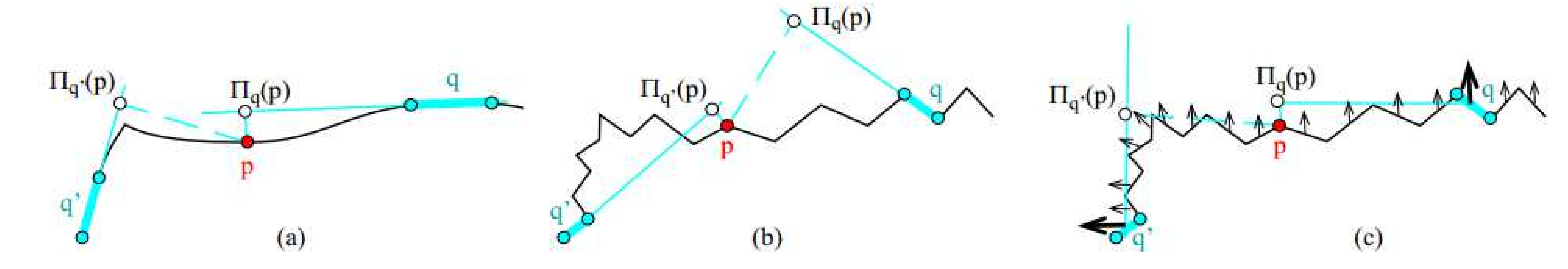}
\caption{(a) A point $p$ is predicted as ${\textstyle \prod_{q}^{} }(p)$ based on the surface at $q$ is the projection of $p$ to the plane tangent to the surface at $q$. As we see the predictions are far away constructed from points across sharp features. (b) Noisy normals results in poor prediction (c) modification after mollification.  Figure reproduced from Thouis et al.\cite{thouis}}
\label{fig:fig28}
\end{figure}

\begin{figure}
\centering
\includegraphics[height=5 cm, width=14 cm]{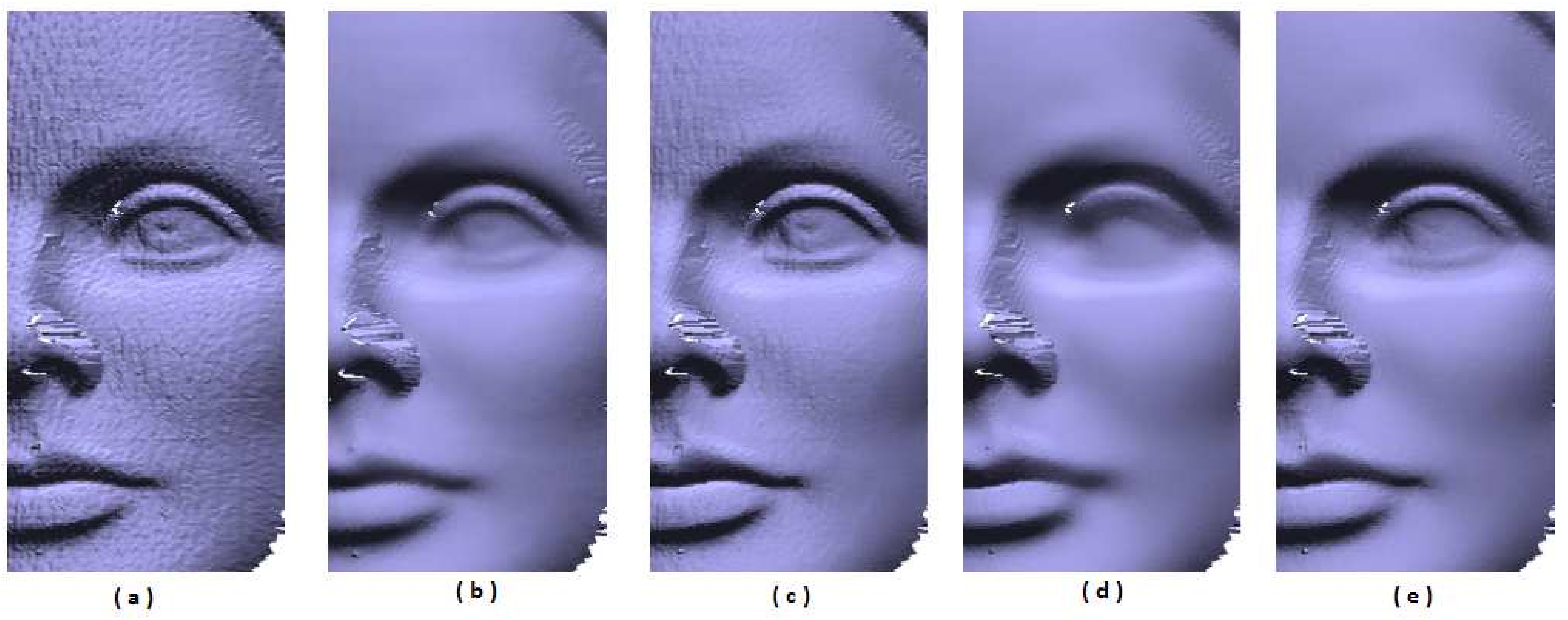}
\caption{(a) Original image with mesh. (b) Isotropic smoothing \cite{desbrun}. (c) approach without mollification. (d) approach with no influence weight $g$. (e) complete approach of Thouis \cite{thouis}.  Figure reproduced from Thouis et al.\cite{thouis}.}
\label{fig:fig29}
\end{figure}

Thouis et al.\cite{thouis} showed a noniterative and robust mesh smoothing technique to remove noise while maintaining the feature using a robust statistical approach and local first order predictor on surface. Predictors have been used based on triangles of mesh on natural tangent planes to the surface. The readers should a the Figure. \ref{fig:fig28} for clear understanding. A new predicted point $p$ is being generated based on the predictor ${\textstyle \prod_{q}^{} }(p)$ from its nearby spatial triangle. The spatial weight $f$ depends on the distance $\left\Vert p-c_q \right\Vert$ between $p$ and the centroid $c_q$ of $q$. The range component $g$ depends on the $\left\Vert  {\textstyle \prod_{q}^{} }(p)-p\right\Vert$ between the prediction and the original source point $p$. Therefore the prediction $p^{'}$ on a surface is given by:
 
\begin{equation}
p^{'} =\frac{1}{k(p)}\sum_{q\epsilon s}^{}{\textstyle \prod_{q}^{} }(p)a_qf(\left\Vert  c_q -p\right\Vert)g(\left\Vert  {\textstyle \prod_{q}^{} }(p)-p \right\Vert),
\end{equation}

where the normalizing factor $k$ is :

\begin{equation}
k(p) =a_qf(\left\Vert  c_q -p\right\Vert)g(\left\Vert  {\textstyle \prod_{q}^{} }(p)-p \right\Vert),
\end{equation}

and $a_q$ is the area of the triangles used to weight for the variations of sampling rate of surface.\\
In Figure \ref{fig:fig28} the normals as shown are more prone to noise than vertices as they are first-order properties of the mesh as shown in Figure \ref{fig:fig28}b. However the estimation can be increased with mollification \cite{huber,murio}, which is achieved by smoothing the normals without changing the positions of the vertices during mollification to preserve the features as few corners might get improperly smoothed by mollification near corners. The results are shown in Figure \ref{fig:fig29}.
\par There are other applications of bilateral filtering which includes video stylization \cite{winnemoller}, demosaicking \cite{bayer}, optical flow  \cite{xiao} etc.

\section{Connection among various edge preserving filters}

There exists a number of existing edge preserving smoothing filters having various contributions in graphics and image processing and among them we find bilateral filter and anisotropic diffusion as the most dominating one and will show their relationships with one another and with other methods as they give similar results.

\begin{figure}
\centering
\includegraphics[height=7 cm, width=15 cm]{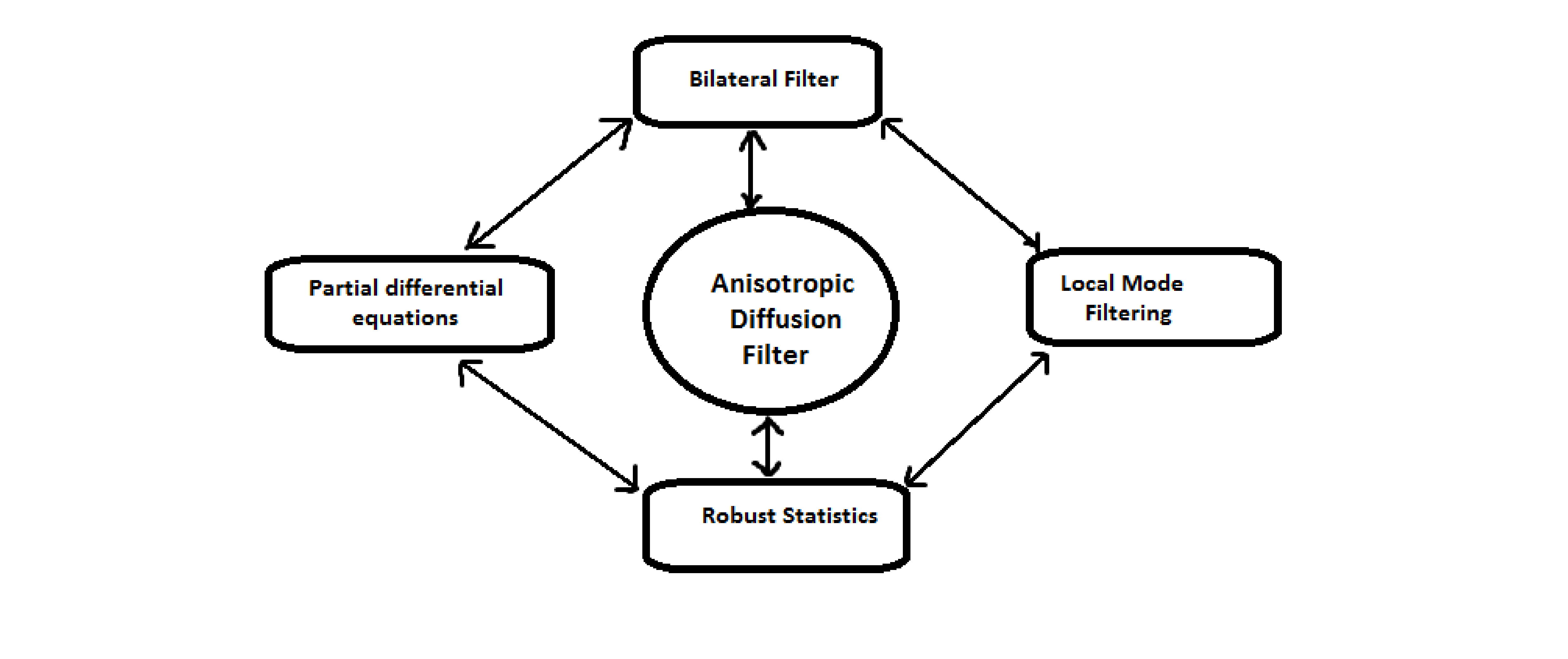}
\caption{Relation among various edge preserving filters.}
\label{fig:fig30}
\end{figure}

The bilateral filter acts as a bridge between anisotropic diffusion filtering and other filtering methods which is shown pictorially as in Figure \ref{fig:fig30}. 

\subsection{ Yaroslavsky filter with bilateral filter}
 
Let us refer to an ingeneral equation as

\begin{equation}
v(i)=u(i)+ n(i)
\end{equation}

where $v(i)$ is the noisy output image, $u(i)$ is the pure ideal image and $n(i)$ is the additive noise.
Yaroslavsky filter
averages the pixels with a similar grey level value and belonging to the spatial neighborhood $B_\rho(x)$ given by

\begin{equation}
YF_{h,\rho}u(x)=\frac{1}{C(x)}\int_{B_\rho(x)}u(y)e^{-\frac{\left | u(y)-u(x) \right |^2}{h^2}}dy,
\label{YF} 
\end{equation}

where $x\epsilon \Omega, C(x)=\int_{B_\rho(x)}e^{-\frac{\left | u(y)-u(x) \right |^2}{h^2}}dy$ is a normalizing factor and $h$ is a filtering parameter.
Later on in 1995 and in 1998 SUSAN filter \cite{smith} and the bilateral filter \cite{tomasi} got introduced  which instead of considering a fixed spatial neighborhood $B_\rho(x)$, weigh the distance 
to the reference pixel $x$. given by

\begin{equation}
BF_{h,\rho}u(x)=\frac{1}{C(x)}\int_{\Omega}u(y)e^{\frac{-\left | y-x \right |^2}{\rho^2}}e^{-\frac{\left | u(y)-u(x) \right |^2}{h^2}}dy,
\label{BF} 
\end{equation}

As per Baudes in 2005 \cite{baudes} there is no as such difference between bilateral filter(BF) and Yaroslavsky filter(YF). Bilateral filter with a square box window is simply an approximation of the Yaroslavsky Filter which restricts the integral to a spatial neighborhood independent of the position $y$. In other words Yaroslavsky filter is a special case of Bilateral filter with a step function as the spatial weight. As a result we don't find the term $\left | y-x \right |^2$ in the Yaroslavsky filter in equation \ref{YF} unlike in \ref{BF}. If the gray level intensity difference is larger than $h$, then both the filters compute the averages of the pixels belonging to the same region of the reference pixel thereby behaves as an edge preserving smoothing filter.

\subsection{Relation of bilateral filter with anisotropic diffusion filter}
There exists a wise link between anisotropic diffusion filtering and bilateral filter and here we would like to focus on the discrete domain computation.
Fortunately it has been observed by D.Barash \cite{barash} that adaptive smoothing serves as a link between anisotropic diffusion and bilateral filtering.\\
\textbf{\emph{Anisotropic diffusion and adaptive filtering}:} 

Give an image $I^t(\overrightarrow{x})$, where $\overrightarrow{x}=(x1,x2)$ are the spatial coordinates, and the iteration of the adaptive filtering yields: 

\begin{equation}
I^{(t+1)}(\overrightarrow{x})=\frac{{\textstyle \sum_{i=-1}^{+1}{\textstyle \sum_{j=-1}^{+1} } I^t(x1+i,x2+j)w^t}}{{\textstyle \sum_{i=-1}^{+1}{\textstyle \sum_{j=-1}^{+1}w^t } }}
\end{equation}

where the convolution mask $w^t$ is defined as :

\begin{equation}
w^t(x1,x2)=exp(-\frac{\left| d^t(x1,x2) \right|^2}{2k^2})
\end{equation}

$k$ the variance of Gaussian kernel, where $d^t(x1,x2)$ is given by the magnitude of the gradient computed as:

\begin{equation}
d^t(x1,x2)=\sqrt{G_{x1}^2 + G_{x2}^2}
\end{equation}

where 

\begin{equation}
(G_{x1},G_{x2})=(\frac{\partial I^t(x1,x2)}{\partial x1},\frac{\partial I^t(x1,x2)}{\partial x2})
\end{equation}

which bears similarity between the convolution kernel of the diffusion coefficient of the anisotropic diffusion.\\
Considering one dimensional signal of $I^t(x)$ and showing the averaging process as follows:

\begin{equation}
I^{t+1}(x)=c_1I^t(x-1)+c_2I^t(x)+c_3I^t(x+1)
\end{equation}

where 
\begin{equation}
c_1+c_2+c_3=1
\end{equation}

which can also be written as,

\begin{equation}
I^{t+1}(x)-I^t(x)=c_1(I^t(x-1)-I^t(x))+c_3(I^t(x+1)-I^t(x))
\end{equation}

replacing $c_1=c_3$ the above equation reduces to:

\begin{equation}
I^{t+1}(x)-I^t(x)=c_1(I^t(x-1)-2I^t(x)+I^t(x+1))
\end{equation}

which can be written as a discrete approximation of the linear diffusion equation as:

\begin{equation}
\frac{\partial I}{\partial t}=c\bigtriangledown^2I
\end{equation}

Considering a second case when the weights are space-dependent:

\begin{equation}
I^{t+1}(x)-I^t(x)=c^t(x+1)(I^t(x+1)-I^t(x))-c^t(x-1)(I^t(x)-I^t(x-1))
\end{equation} 

which finds similarity with the implementation with the anisotropic diffusion equation by Perona and Malick \cite{perona}

\begin{equation}
\frac{\partial I}{\partial t}=\bigtriangledown(c(x1,x2)\bigtriangledown I)
\end{equation}

where 

\begin{equation}
c(x_1,x_2)=g(\left\Vert \bigtriangledown I(x_1,x_2) \right\Vert)
\end{equation}

and $\left\Vert \bigtriangledown I\right\Vert$ is the magnitude of the gradient and $g(\left\Vert  \bigtriangledown I\right\Vert)$ is an edge stopping function satisfying every parameters.
Thus a link between anisotropic diffusion(60) and adaptive 
filtering(50) has been established.\\

\textbf{\emph{Bilateral filter and Adaptive filtering.}}

The discrete version of the bilateral filter is given below:

\begin{equation}
\overrightarrow{F}^{t+1}(\overrightarrow{x})=\frac{{\textstyle \sum_{i=-S}^{+S}{\textstyle \sum_{j=-S}^{+S}\overrightarrow{I^t}(x_1+i,x_2+j)w^t  } }}{{\textstyle \sum_{i=-S}^{+S}{\textstyle \sum_{j=-S}^{+S}w^t } }}
\end{equation}

where the weight is given by:

\begin{equation}
w^t(\overrightarrow{x},\overrightarrow{\xi})=exp(\frac{-(\overrightarrow{\xi}-\overrightarrow{x})}{2\sigma_D^2})exp(\frac{-(I(\overrightarrow{\xi})-I(\overrightarrow{x}))^2}{2\sigma_R^2})
\end{equation}

where $S$ is the kernel size of the filter and the generalized intensity consisting of domain and range components is given by:

\begin{equation}
\widehat{\overrightarrow{I}}\equiv \left\{ \frac{\overrightarrow{I}(\overrightarrow{x})}{\sigma_R},\frac{\overrightarrow{x}}{\sigma_D} \right\}
\end{equation}

\begin{eqnarray}
w^t(\overrightarrow{x})=exp(-\frac{1}{2}\left| \widehat{\overrightarrow{I}}(\overrightarrow{\xi})-\widehat{\overrightarrow{I}}(\overrightarrow{x}) \right|^2) \nonumber\\
=exp(-\frac{1}{2}\left| \left\{\frac{\overrightarrow{I}(\overrightarrow{\xi})}{\sigma_R},\frac{\overrightarrow{\xi}}{\sigma_D} \right\}-\left\{\frac{\overrightarrow{I}(\overrightarrow{x})}{\sigma_R},\frac{\overrightarrow{x}}{\sigma_D} \right\}\right|^2) \nonumber \\
=exp(-\frac{1}{2}\left| \left\{\frac{\overrightarrow{I}(\overrightarrow{\xi})-\overrightarrow{I}(\overrightarrow{x})}{\sigma_R},\frac{\overrightarrow{\xi}-\overrightarrow{x}}{\sigma_D} \right\}\right|^2) \nonumber\\
=exp(-\frac{1}{2} (\frac{(\overrightarrow{I}(\overrightarrow{\xi})-\overrightarrow{I}(\overrightarrow{x}))^2}{\sigma_R}+\frac{(\overrightarrow{\xi}-\overrightarrow{x})^2}{\sigma_D} ) )\nonumber\\
=exp(-\frac{(\overrightarrow{\xi}-\overrightarrow{x})^2}{2\sigma_D^2} ) 
exp(-\frac{(\overrightarrow{I}(\overrightarrow{\xi})-\overrightarrow{I}(\overrightarrow{x}))^2}{2\sigma_R^2})
\end{eqnarray}

Thus it can be seen from equation 65 that it contains two components  namely the domain and range kernel weights respectively which proves the link between bilateral filtering and adaptive filtering.
Thus it can be concluded that since there is a link between bilateral filter and anisotropic diffusion filter via adaptive smoothing filter so knowing one the other can be derived.

\subsection{Bilateral filter with local mode filtering and robust mode filtering}

\begin{figure}
\centering
\includegraphics[height=5.7 cm, width=15 cm]{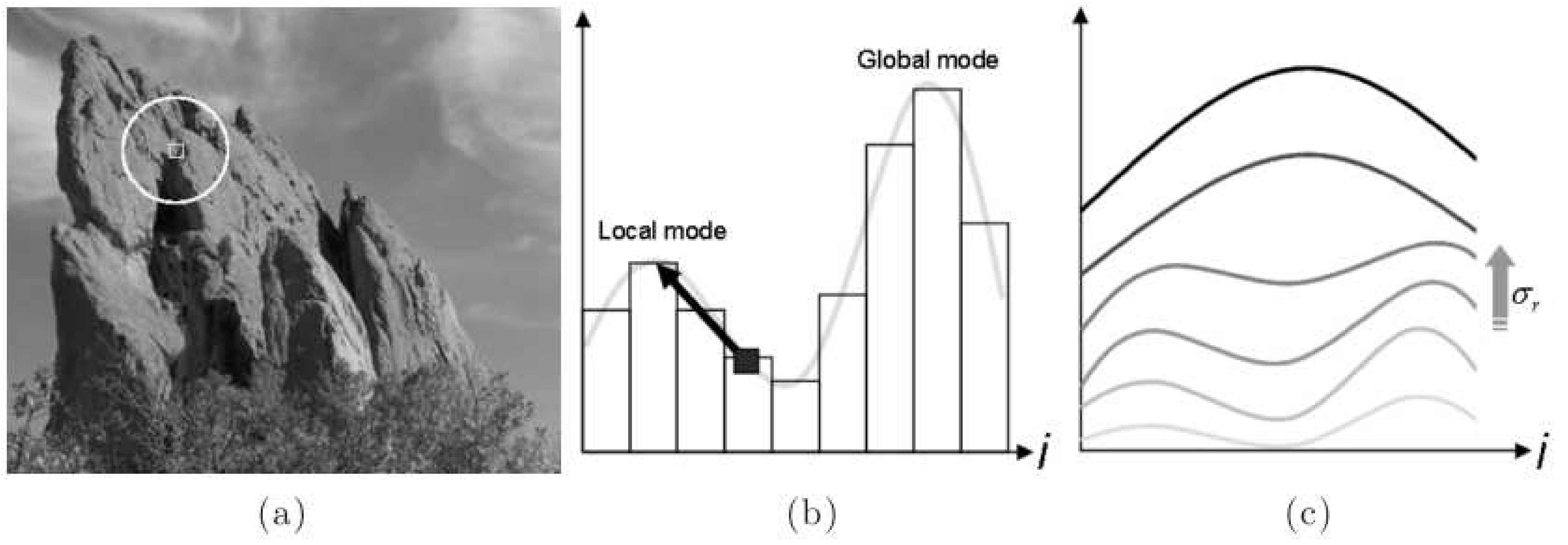}
\caption{(a) A selected window and its neighborhood meant for histogram calculation. (b) As per the figure each pixel tends to shift towards the maximum of the local mode. (c) Curve shows the smoothing effect of the local histogram upon change in the range parameter. concept of local mode filtering reproduced from \cite{local}. }
\label{fig:fig31}
\end{figure}

Sylvain et al.\cite{parissignal} have observed that starting from local mode filtering, the bilateral filtering can be derived. It can also be achieved while trying to find a solution of a robust minimization problem. Here we like to summarize it. The local mode is basically chosen unlike the global mode(shown in Figure \ref{fig:fig31}c.) to preserve the fine details of an image. 

Given a grayscale image $I$ the local mode filtering can be defined as:

\begin{equation}
H^1(i)=\sum_{q\epsilon S}^{}\delta(I_q-i), \forall i\epsilon R,
\end{equation}

such that $I:\Omega\rightarrow R,$. $\delta$ is the Dirac delta function. Since the local mode filtering also depends upon the smoothness parameter of the histogram, so it can be defined as,

\begin{equation}
H^2(i,\sigma_r)=H^1\otimes G_{\sigma_r}(i)=\sum_{q\epsilon S}^{}G_{\sigma_r}(I_q-i),
\end{equation}

where $\sigma_r$ is the intensity range within which the smoothing is done. Now to define a histogram locally i.e around a given position $p$ a spatial Gaussian kernel needs to be applied around the point $p$ to get:

\begin{equation}
H^3(p,i,\sigma_s,\sigma_r)=\sum_{q\epsilon S}^{}G_{\sigma_s}(\left\Vert  p-q\right\Vert)G_{\sigma_r}(I_q-i),
\end{equation}
where $\sigma_s$ denotes the spatial neighborhood centred around pixel position $p$.

As all the pixels within the local neighborhood tends to move towards the maximum local range centered at $I_p$ so it is required to equate the equation to zero as shown:

\begin{equation}
\frac{\partial H^3}{\partial i}(p,i,\sigma_r,\sigma_s)\mid_{i=I_p}=0.
\end{equation}

The equation 69 equates to be:

\begin{equation}
\sum_{q\epsilon S}^{}(I_q-i)G_{\sigma_s}(\left\Vert p-q \right\Vert)G_{\sigma_r}(I_q-i)=0,
\end{equation}

where $I_p$ needs to verify the following implicit equation, equating $I_p=i$ where $i$ is:

\begin{equation}
=\frac{\sum_{q\epsilon S}^{}G_{\sigma_s}(\left\Vert p-q \right\Vert)G_{\sigma_r}(I_q-i)I_q}{\sum_{q\epsilon S}^{}G_{\sigma_s}(\left\Vert p-q \right\Vert)G_{\sigma_r}(I_q-i)}
\end{equation}

Since the local mode filtering
is an iterative procedure which converges to the closest highest mode of the local histogram, one needs to propose an iterative scheme to solve the implicit equation 71 to get:

\begin{equation}
I_p^{t+1}=\frac{\sum_{q\epsilon S}^{}G_{\sigma_s}(\left\Vert p-q \right\Vert)G_{\sigma_r}(I_q^t-I_p^t)I_q^t}{\sum_{q\epsilon S}^{}G_{\sigma_s}(\left\Vert p-q \right\Vert)G_{\sigma_r}(I_q^t-I_p^t)}
\end{equation}

for all $p$.

So interestingly it has been observed that their is a good resemblance between bilateral filter equation with that of Equation 72.
\par 
It has been observed that the bilateral filtering correlate to a gradient descent of a robust minimization problem. 
\emph{The problem definition}: Input is a noisy image $I^n$.\\
It is required to minimize the discrete equation:

\begin{equation}
\underset{I}{min}\sum_{p\epsilon S}^{}((I_p-I_p^n)^2+\sum_{q\epsilon N(p)}\rho(I_q-I_p)),
\end{equation}

where $p$ and $q$ are pixel positions and $N(p)$ is the neighborhood of $p$, $\rho$ is the weighting function.
The second term of equation 73 is the regularization term which is sensitive to the intensity difference among neighboring pixels, influenced by the function $\rho$.
This regularization term helps to find out the relationship with bilateral filter. To do that a new reweighed regularization term has been introduced to make the minimization problem take the form:

\begin{equation}
\underset{I}{min}\sum_{p\epsilon S}^{}\sum_{q\epsilon N(p)}G_{\sigma_s}(\left\Vert q-p \right\Vert)\rho(I_q-I_p)),
\end{equation}

To minimize equation 74 the following iteration is done:

\begin{equation}
I_p^{t+1}=I_p^t+ \frac{\lambda}{\left|  N(p)\right|}\sum_{q\epsilon N(p)}G_{\sigma_s}(\left\Vert q-p \right\Vert)\rho^{'}(I_q^{t}-I_p^{t})),
\end{equation}

By substituting $\rho(s)=1-G_{\sigma_r}(s)$ the equation becomes:

\begin{equation}
I_p^{t+1}=I_p^t+ \frac{\lambda}{\left|  N(p)\right|}\sum_{q\epsilon N(p)}G_{\sigma_s}(\left\Vert q-p \right\Vert)G_{\sigma_r}(I_q^{t}-I_p^{t})(I_q^{t}-I_p^{t}),
\end{equation}

which is similar to the bilateral filter expression, leads to the weighted average as shown below.

\begin{equation}
I_p^{t+1}=\frac{{\textstyle \sum_{q}^{}G_{\sigma_s}(\left\Vert q-p \right\Vert)G_{\sigma_r}(I_q^t-I_p^t)I_q^t }}{{\textstyle \sum_{q}^{}G_{\sigma_s}(\left\Vert q-p \right\Vert)G_{\sigma_r}(I_q^t-I_p^t)}}
\end{equation}

This form can be used for solving the minimization problem.
Thus equation 76 and 77 clearly reveals the fact that bilateral filter is a robust filter.

\subsection{Robust Statistics and anisotropic diffusion}

Here we like to revisit the contribution of M J. black et al.\cite{black} who, with the very simple formulation showed the link between anisotropic diffusion and robust statistics.  
Equation 78 shows the continuous anisotropic diffusion filter form. 
\begin{equation}
\frac{\partial I(x,y,t)}{\partial t}=div[g(\left\Vert \bigtriangledown I \right\Vert)\bigtriangledown I]
\end{equation}

Similarly eqn. 79 shows the continuous form of the robust estimation problem which focuses on minimizing the equation as discussed in the previous section.

\begin{equation}
\underset{I}{min}\int_{\Omega}\rho(\left\Vert \bigtriangledown I \right\Vert)d\Omega
\end{equation}

where $\Omega$ is the image domain. Minimization of eqn. 79 is done via gradient descent using calculus of variations as shown:

\begin{equation}
\frac{\partial I(x,y,t)}{\partial t}=div[\rho^{'}(\left\Vert \bigtriangledown I \right\Vert)\frac{\bigtriangledown I}{\left\Vert \bigtriangledown I \right\Vert}].
\end{equation}

By substituting 

\begin{equation}
g(x)=\frac{\rho^{'}(x)}{x}
\end{equation}

the relation between robust estimation in eqn. 79 and anisotropic diffusion in eqn. 78 used for image reconstruction  is obtained.

\subsection{Partial Differential Equation and Anisotropic Diffusion Algorithm}

In many imaging applications it has beeen observed that unknown data are related to observed data through a linear equation 

\begin{equation}
p=Rf+v
\end{equation}
where $v$ is the white Gaussian noise, $R$ is a linear operator defined on $L^2(R^2)$, considering the general problem of recovering unknown data $f$ from noisy observed data $p$.
Teboul et. al \cite{teboul} considered a  minimisation function in order to estimate $f$ from $p$ defined as

\begin{equation}
J(f)=\int_{\Omega}\left | f(x)-p(x) \right |^2dx+\lambda^2\int_{\Omega}\varphi[\left | \bigtriangledown f(x) \right |] dx
\end{equation}

where $\Omega$ is an open bound set in $R^2$, $\left | \bigtriangledown f(x) \right |$ is the modulus of the gradient of $f(x)$, $\Lambda$ is a regularization
parameter which balances the influence between the two terms of equation 81.
Minima of (81), if they exist, formally verify the Euler Lagrange equation given by

\begin{eqnarray}
[f(x)-p(x)]-\lambda^2div{\frac{\varphi'[\left | \bigtriangledown f(x) \right |]}{2\left | \bigtriangledown f(x) \right |}} \bigtriangledown f(x)=0, x\epsilon \Omega \nonumber \\ 
\frac{\partial f}{\partial n}|_{\partial \Omega} =0
\end{eqnarray}

where $n$ denotes the vector normal to the boundary $d\Omega$ of $\Omega$, div is the divergence operator given by

\begin{equation}
div(u)=\sum_{k=1}^{2 or 3}\frac{\partial u_k }{\partial x_k}
\end{equation}

The nonlinear equation (82) has a close resemblance to the anisotropic diffusion filter \cite{perona}.

Equation 82 can be solved by a gradient-descent method given by:

\begin{eqnarray}
\frac{\partial f(t,x))}{\partial t}=[p(x)-f(t,x)]+\lambda^2div{\frac{\varphi'[\left | \bigtriangledown f(t,x) \right |]}{2\left | \bigtriangledown f(t,x) \right |}} \bigtriangledown f(t,x)=0 \nonumber \\ 
\frac{\partial f}{\partial n}|_{\partial \Omega} =0
\end{eqnarray}

where $t$ is a scale parameter.
The nonlinear PDE (84) corresponds to the anisotropic diffusion equation of Perona and Malik \cite{perona}.

\section{Recent extensions/variations of bilateral filter}

There are several variants and extensions of the most extensively used bilateral filter. To name a few we like to discuss upon them in the next sequel.\\

\subsection{The Trilateral filter}

\begin{figure}
\centering
\includegraphics[height=5 cm, width=15.5 cm]{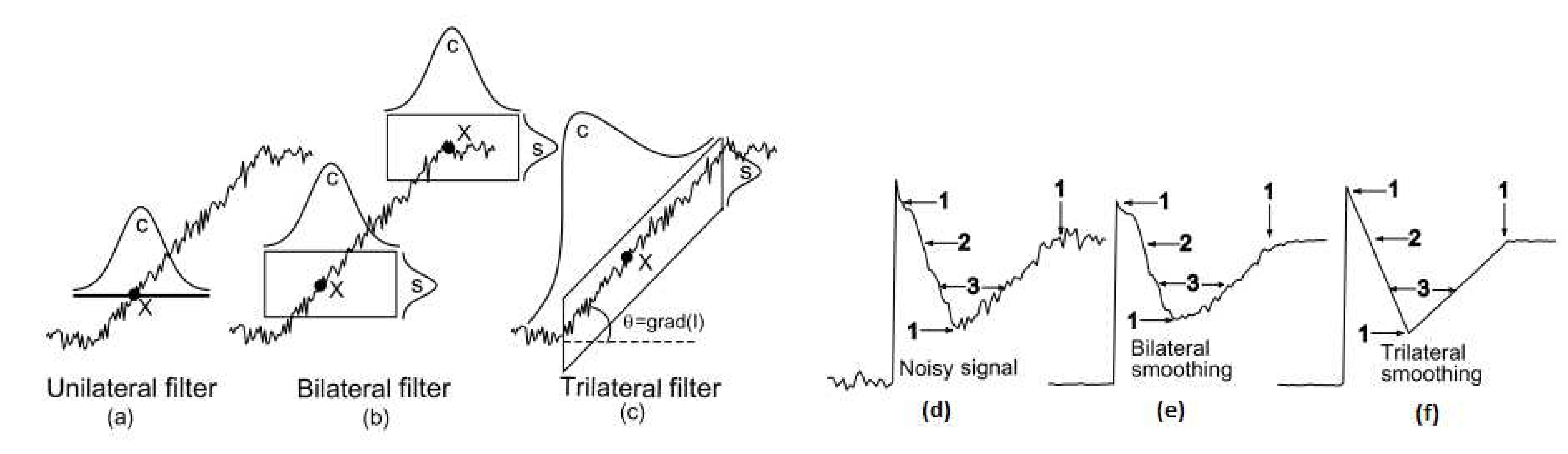}
\caption{(a,b and c) shows the unilateral, bilateral and trilateral filter range for one scan line of an image. Now given a (d) noisy signal (e) shows the bilateral filter smoothes the sharp corners(e.1) and averages the high gradient regions poorly(e.2) , (f) the trilateral filter smoothes corners(f.1) and high gradient regions(f.1) well, where 1,2 and 3 denotes ridge and valley like edges, high-gradient regions, and similar intensities in disjoint regions respectively. Figure reproduced from P.Choudhury \cite{prasun}.}
\label{fig:fig32}
\end{figure}

This trilateral filter as introduced by P.Choudhury et al.\cite{prasun} in the year 2003 is another variant and an extension of the bilateral filter for edge preserving smoothing operations. But it is meant for handling $N-$dimensional signals in multimedia applications. Unlike their classical counterparts like anisotropic diffusion and bilateral filter, it is much robust for noise reduction and better outlier rejection in high-gradient regions.\\
Applying the bilateral filter in regions of sharp change in gradient and high gradient areas degrade the smoothing capabilities of the bilateral filter as the rectangular filter extent encloses pixels spanning the peak of the ridge or valley as shown in Figure \ref{fig:fig32}. As a result the filter blends these intensities and results in a blunt feature. As shown in Figure \ref{fig:fig32} the spatial extent of the filter can accommodate only a narrow portion of the input signal. These problems have been addressed by Choudhury et al.\cite{prasun} by combining modified bilateral filters with a pyramid-based method to limit filter extent. \\
Initially they have estimated the slopes of the intensity gradient using bilateral filter followed by tilting the filter extent by an angle $\theta$ pivoting the center point at $(x,I(x))$ to cover a wide area of usable domain. 
. Its tilting vector $G_{\theta}(x)$
should average together closely related neighborhood gradients ignoring nearby strongly dissimilar gradient outliers thereby assists in restoring the effectiveness of the spatial filter term. The tilting vector is given by:

\begin{equation}
G_{\theta}(x)=\frac{1}{k_{\theta}(x)}\int_\infty^\infty \bigtriangledown I_{in}(x+\zeta)c(\zeta)s(\left\Vert \bigtriangledown I_{in}(x+\zeta)-\bigtriangledown I_{in} (x)\right\Vert)d\zeta
\end{equation}

where

\begin{equation}
k_{\theta}(x)=\int_\infty^\infty c(\zeta)s(\left\Vert \bigtriangledown I_{in}(x+\zeta)-\bigtriangledown I_{in} (x)\right\Vert)d\zeta.
\end{equation}

\begin{figure}
\centering
\includegraphics[height=8 cm, width=16 cm]{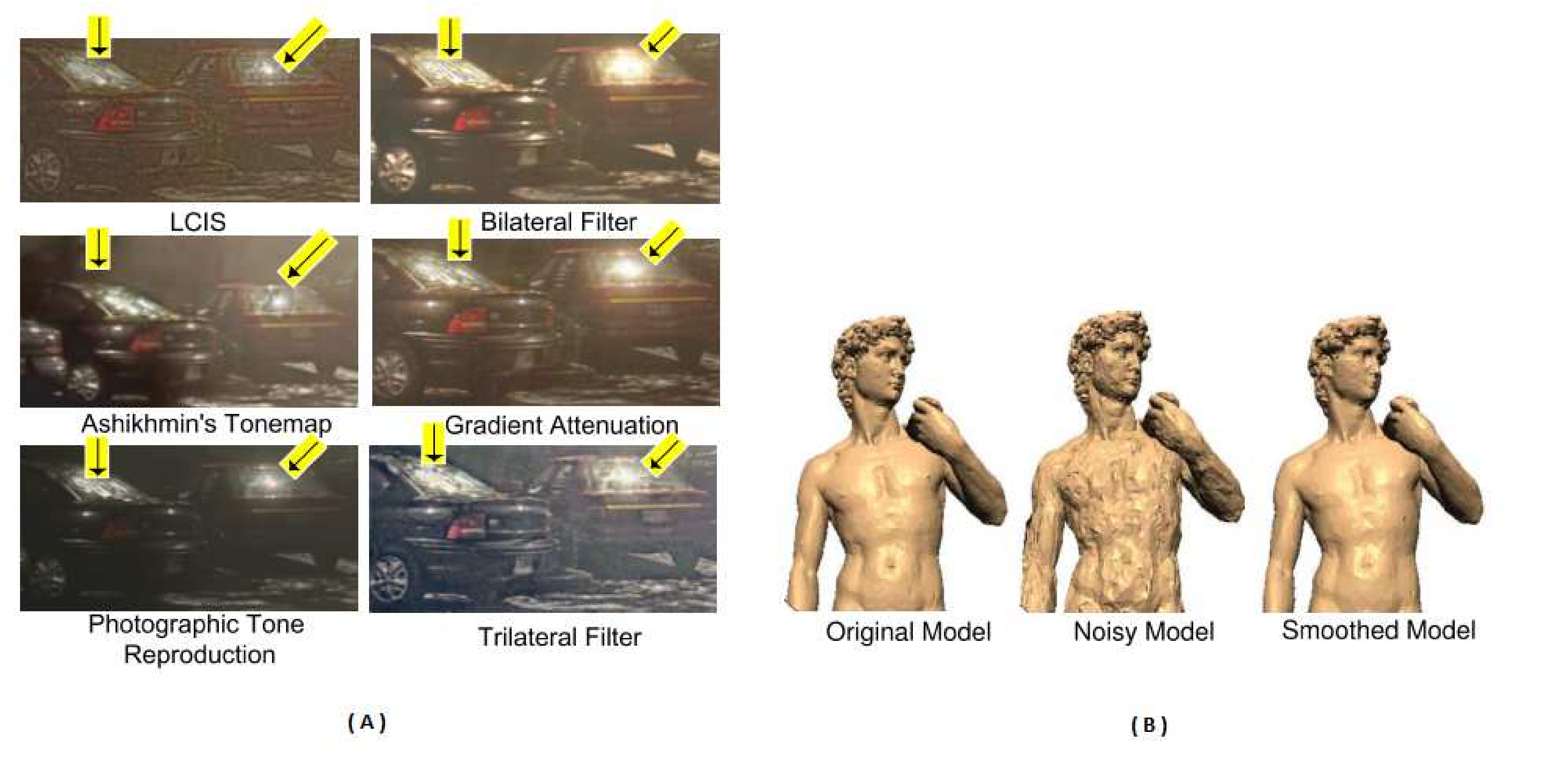}
\caption{(A) In comparison with the bilateral filter, trilateral filtering limits blooming of sharp specular highlights and performs similar to the gradient attenuation method. (B) The single pass trilateral filter removes most visible corruptions caused by additive Gaussian noise in both vertex positions and normals. Figure reproduced from P.Choudhury \cite{prasun}.}
\label{fig:fig33}
\end{figure}

\begin{figure}
\centering
\includegraphics[height=4.5 cm, width=15 cm]{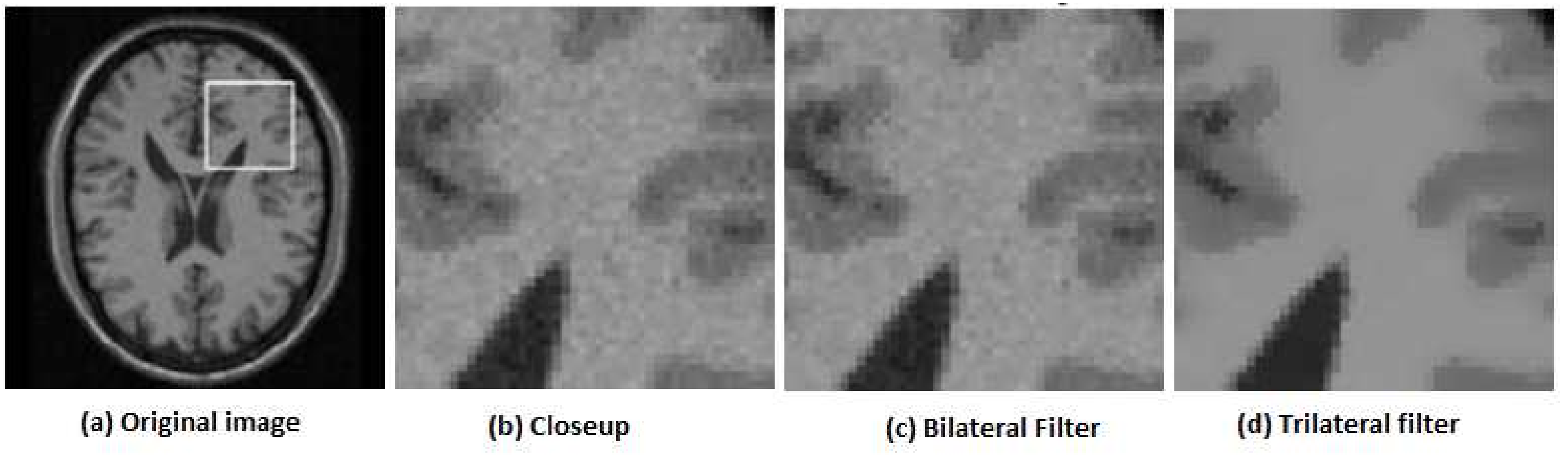}
\caption{MRI image (a) An original slice image. (b) magnified view of the slice. (c) After applying bilateral filter on the slice region. (d) After applying trilateral filter. Figure reproduced from Wong et.al. \cite{wong}.}
\label{fig:fig34}
\end{figure}

$c()$ denotes the domain of filter window, $s()$ the range component. Instead of computing a range weight $s()$
for neighboring $I(x+\zeta)$ by measuring its closeness to the center point value $I(x)$ , instead its closeness
to a plane through $I(x)$ , which acts
as a \emph{centerline} for the filter window as shown in Figure \ref{fig:fig32}c. It takes only one user set parameter, filters a signal in a single pass(thus non-iterative) as needed by most PDE based methods. As already mentioned it can handle $N-$dimensional signal and provides much better visual quality in applications like texture preserving in contrast reduction problem in digital photography and while denoising polygonal meshes. As shown in Figure \ref{fig:fig33}A and B. As mentioned by W.C.K Wong et al.\cite{wongtri} in their benchmark paper on medical image processing, that a narrow spatial window, say, 3 pixels in each dimension, should be used in order to avoid over-smoothing structures of sizes comparable to the image resolutions as sharp ridges and gutters which are commonly found in biomedical images, such as nested vessels in digital subtraction angiography (DSA) and 3D angiography, folded gray and white matters in brain MR imaging. Wide spatial window generally over-smooth sharp ridges, gutters and fine textures in an image \cite{prasun}.  So a trade-off is necessary to strike a balance between the size of the spatial window and the number of iterations needed to be performed in bilateral filtering. Their results have been shown in Figure \ref{fig:fig34}.\\ T.Vaudrey et al.\cite{vaudrey} in the year 2009 have shown that the trilateral filtering can further be speeded up  several orders of magnitude (100-1000 times faster; from hours down to seconds) by the use of LUT(look Up Table) and by truncating the LUT to a user specified numerical accuracy. 
Referring to equation 82, $c()$ (the spatial weight) can be pre-calculated using a look-up-table (LUT) as t depends on the domain kernel standard deviation (say $\sigma_1$) and 
$m$ (say the kernel size) and $a$(offset from central pixel).
\emph{N.B}: size of the used filter kernel $2m+1$
$m$ is the half kernel size and is $n-$dimensional.
As this function is Gaussian, the LUT can be computed as:

\begin{equation}
c(i)=exp(\frac{-\left\Vert i \right\Vert^2}{2\sigma_1^2})
\end{equation}

where $0\leq i \leq m.$ $s()$ cannot use a look up table since it is depends on the local neighborhood intensity. 
Numerical approximation needs to be done. The LUT based approach is a smart truncation of the kernel to a defined accuracy $\varepsilon$, where $0<\varepsilon<1$. If any value below $\varepsilon$ needs to be truncated then all the above it should be used given by,

\begin{equation}
\varepsilon < exp(\frac{-\left\Vert i \right\Vert^2}{2\sigma_1^2})
\end{equation}

which leads to $\left\Vert  i\right\Vert\leq \sigma_1\sqrt{-2ln(\varepsilon)}=T$

where $T$ is the threshold. This approach is not much beneficial to bilateral filtering but upon applying it to trilateral filtering, it reduces the number of computing equations drastically. This implementation is open to massive parallel processing (every pixel is independent within the iteration of trilateral filter) and opens up a huge opportunity for GPU (Graphics Processing Unit) based implementation.

\subsection{Joint and dual bilateral filter}

\begin{figure}
\centering
\includegraphics[height=4.5 cm, width=15 cm]{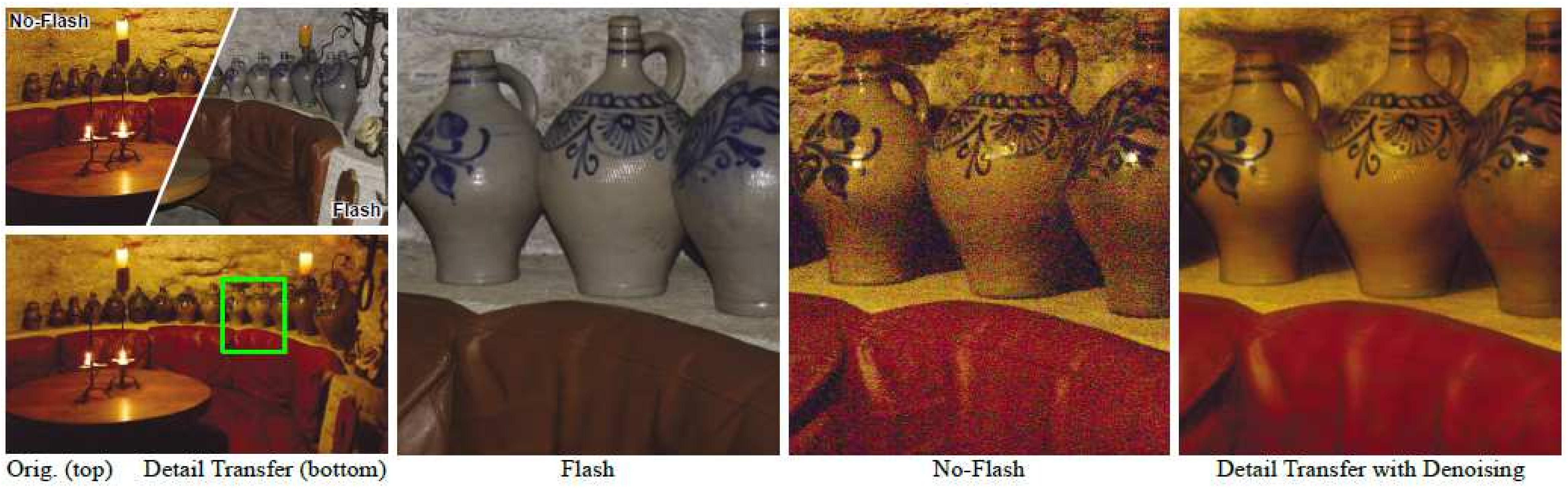}
\caption{The pic shows candlelit settings in flash and noflash environment. A particular area region of the image is magnified as shown and the details of the image  has been extracted with denoising as shown Figure reproduced from G. Petschnigg et al.\cite{pets}.}
\label{fig:fig35}
\end{figure}

This filter commonly known as cross/joint bilateral filter is just a variant of the classical bilateral filter as introduced by Eisemann et al.\cite{eisemann} and Petschnigg et al.\cite{pets}.  
\emph{What it does:} Given two input images $I$ and $E$, the joint bilateral filter smooths $I$, while preserving the edge of the second image $E$. In other words the range component is computed using the image $E$ given by:

\begin{equation}
JBF[I,E]_p=\frac{1}{W_p}\sum_{q \epsilon S}^{}G_{\sigma_s}(\left\Vert p-q \right\Vert)G_{\sigma_r}(E_p-E_q)I_q,
\end{equation}

with 
\begin{equation}
W_p=\sum_{q \epsilon S}^{}G_{\sigma_s}(\left\Vert p-q \right\Vert)G_{\sigma_r}(E_p-E_q)
\end{equation}

and $W_p$ defines the normalization weight, $G_{\sigma_s}$ and $G_{\sigma_r}$ denotes the spatial and range kernel functions respectively.
Figure \ref{fig:fig26}, \ref{fig:fig27} and \ref{fig:fig35} demonstrates the behaviour.\\

\textbf{\emph{Dual Bilateral filter:}}

This is another variant of bilateral filter as introduced by Bennet et al.\cite{bennett}. It takes two images as input $I$ and $J$ as input and both the input images are used to define the edges whereas he previous joint bilateral filter used only one of the flashed image to extract out the edge information.    
It is defined by:

\begin{equation}
DBF[I]_p=\frac{1}{W_p}\sum_{q \epsilon S}^{}G_{\sigma_s}(\left\Vert p-q \right\Vert)G_{\sigma_I}(I_p-I_q)G_{\sigma_J}(J_p-J_q)I_q,
\end{equation}

The merit of the above approach is that there remains very less probability to miss the edge information present in an image of various spectrum and in light or no-light condition.

\subsection{Nonlocal-means}

\begin{figure}
\centering
\includegraphics[height=4.6 cm, width=6 cm]{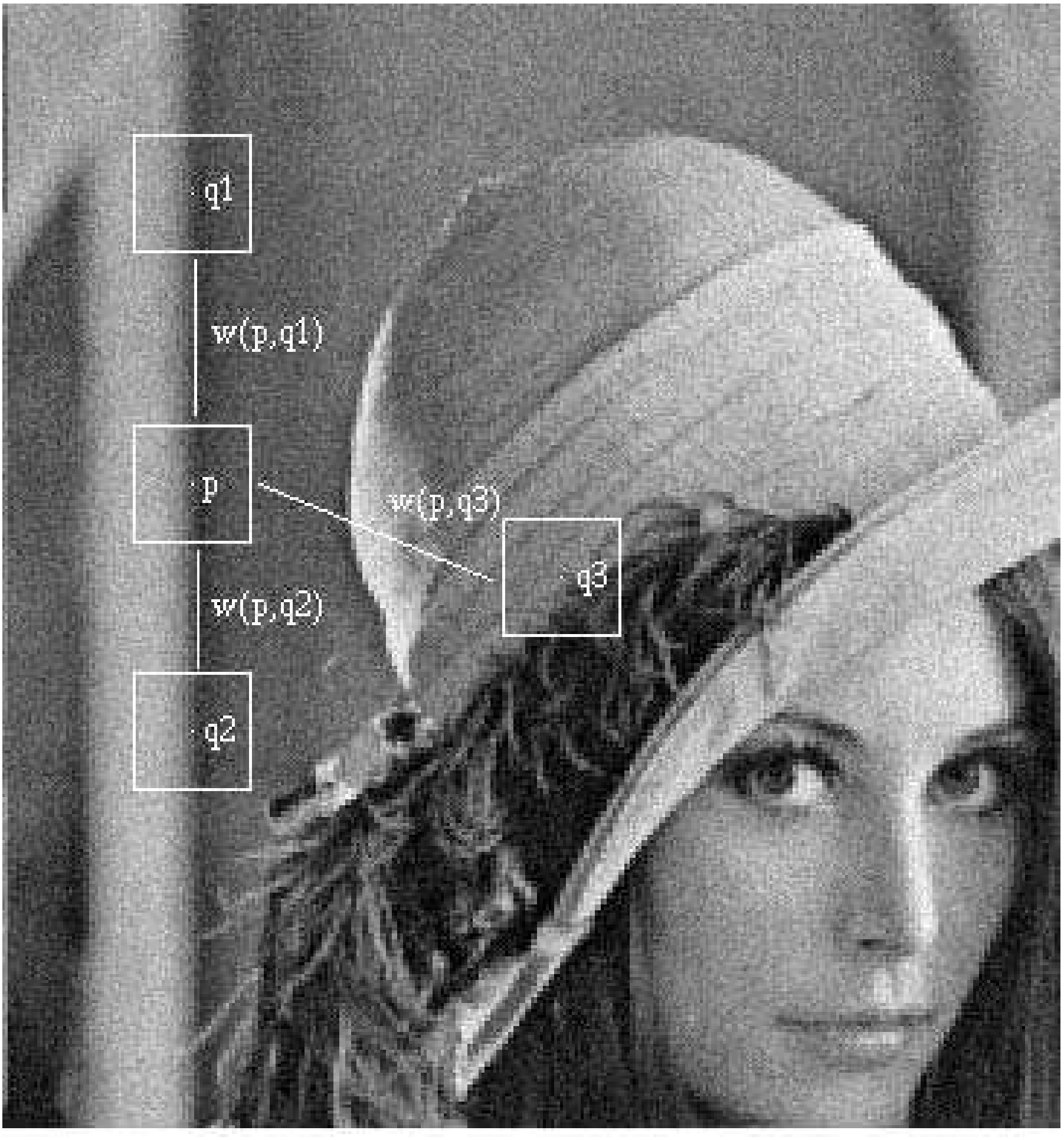}
\caption{Scheme of NL-means strategy. Similar
pixel neighborhoods give a large weight, w(p,q1)
and w(p,q2), while much different neighborhoods
give a small weight w(p,q3). Figure reproduced from Antoni Buades et al.\cite{buades}.}
\label{fig:fig36}
\end{figure}

For each pixel $p$ define a small, simple fixed size neighborhood 
as shown in Fig.\ref{fig:fig36}.
Secondly a vector $V_p$ is defined which  contains a set of neighbouring pixel values. i.e $V_p$=[3 5 1 7 8 3...]. Similar pixels $p$ and $q1$ have small vector distance(weight) given by: $w(p,q1)=\left\Vert V_p-V_{q1} \right\Vert^2$ and $w(p,q2)=\left\Vert V_p-V_{q2} \right\Vert^2$. However  dissimilar pixels such as $p$ and $q3$ have large vector distances given by $w(p,q3)=\left\Vert V_p-V_{q3} \right\Vert^2$.

Therefore the non local means filter NL-means is given by

\begin{equation}
NLM[I]_p=\frac{1}{W_p}\sum_{q \epsilon S}G_{\sigma_r}\left\Vert V_p^\rho-V_q^\rho \right\Vert^2I_q,
\end{equation}

where $W_p=\sum_{q \epsilon S}G_{\sigma_r}\left\Vert V_p^\rho-V_q^\rho \right\Vert^2$ is the normalizing factor.
Vector distance to $p$ sets weight for each pixel $q_i$.

Results shows a) is the noisy source image, b) is the figure obtained after gaussian filtering resulting in low noise and low detail. c) is the figure obtained after anisotropic diffusion filtering with some stairsteps as shown. d) is obtained after bilateral filtering which shows more clear picture than the previous ones but still some stairsteps remained which is prominent near the boundary edges. e) is obtained after applying the variant the NL-Means which exhibits sharpness, low noise and few artifacts.

 \begin{figure}
\centering
\includegraphics[height=4.2 cm, width=15 cm]{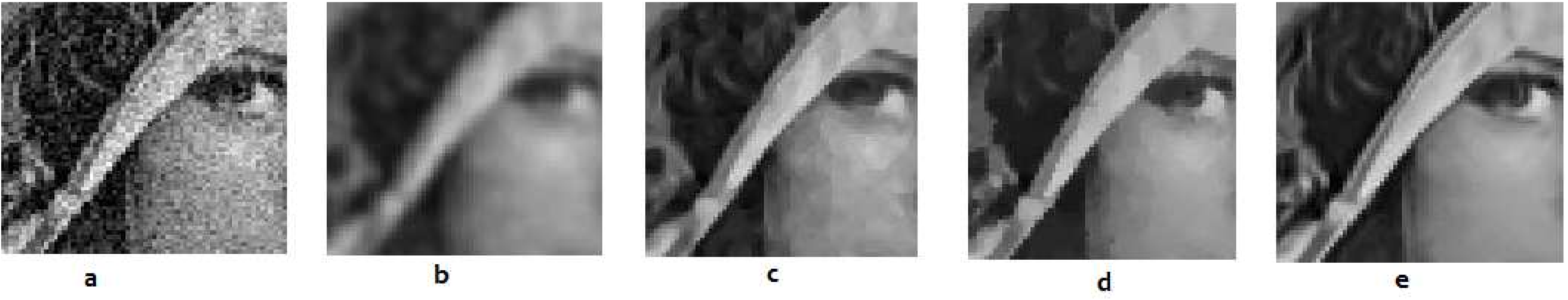}
\caption{a) Noisy input image. b) Gaussian smoothing. c) Anisotropic diffusion filter. d) Bilateral filter. e) NL-Means filtering. Figure reproduced from Antoni Buades et al.\cite{buades}.}
\label{fig:fig37}
\end{figure}

\subsection{Guided filter}

Kaiming in their paper \cite{kaiming} proposed a novel explicit image filter called guided filter. Derived from a local linear model, the guided filter computes the filtering output by considering the content of a guidance image, which can be the input image itself or another different image. The guided filter can be used as an edge-preserving smoothing operator like the popular bilateral filter \cite{tomasi}, but it has better behaviors near edges. Some recent implementations of the bilateral filter \cite{porikli},\cite{parissignal},\cite{yang} have employed quantization techniques to accelerate the execution speed but at the cost of accuracy. The authors have shown unlike bilateral filter 
it does not suffer from the gradient
reversal artifacts as shown in Fig.\ref{fig:fig37}.
It is defined as 
\begin{equation}
q_i=a_kI_i+b_k, \forall i \epsilon w_k
\end{equation}

where ($a_k, b_k$) are some linear coefficients assumed to be constant in $w_k$. The filtering output is $q$ and $I$ is the guided image. In order to determine the linear coefficients ($a_k, b_k$) constraints are needed from the filtering input $p$.
The output $q$ has been modeled from the input $p$ by subtracting some unwanted components $n$ like noise/textures as shown in fig.\ref{fig:fig36}.
\begin{equation}
q_i=p_i - n_i
\end{equation}

\begin{figure}
\centering
\includegraphics[height=4.3 cm, width=15 cm]{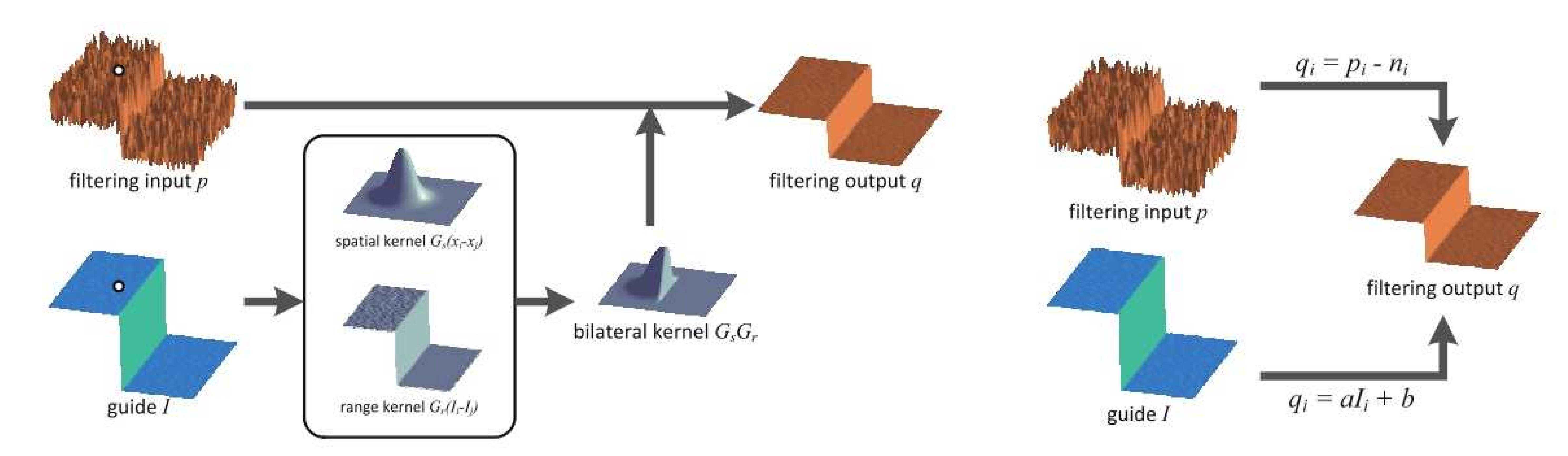}
\caption{Illustrations of the bilateral filtering process (left) and the guided filtering process (right). Figure reproduced from Kaiming He et al.\cite{kaiming}.}
\label{fig:fig38}
\end{figure}

The cost function in the window $w_k$ is minimized as

\begin{equation}
E(a_k,b_k)= \sum_{i\epsilon w_k}^{}{}((a_kI_k+b_k-p_i)^2+\epsilon a_k^2).
\end{equation}

were $\epsilon$ is a regularization parameter penalizing large $a_k$.
The solution of $a_k$ and $b_k$ has been computed as 

\begin{equation}
a_k= \frac{\frac{1}{\left | w \right |}\sum_{i\epsilon w_k I_ip_i-\mu_k\bar{p_k}}^{}}{\sigma_k^2 + \epsilon}
\end{equation}

\begin{equation}
b_k=\bar{p_k}-a_k\mu_k
\end{equation}

where $\mu_k$ and $\sigma_k^2$ are mean and variance of $I$ in $w_k$. Now $a_k$ and $b_k$ being solved it can be used to compute the filter output $q$ in equation 89.
Besides the guided filter has also got some limitations like it may exhibit halos near some edges. \emph{Halos} refer to the artifacts of unwanted smoothing of edges.
It has been suggested by the authors to apply the guided filter in any situation when the bilateral filter works well. The guided filter is much faster and sometimes (though not always) works even better.

\begin{figure}
\centering
\includegraphics[height=4.5 cm, width=10 cm]{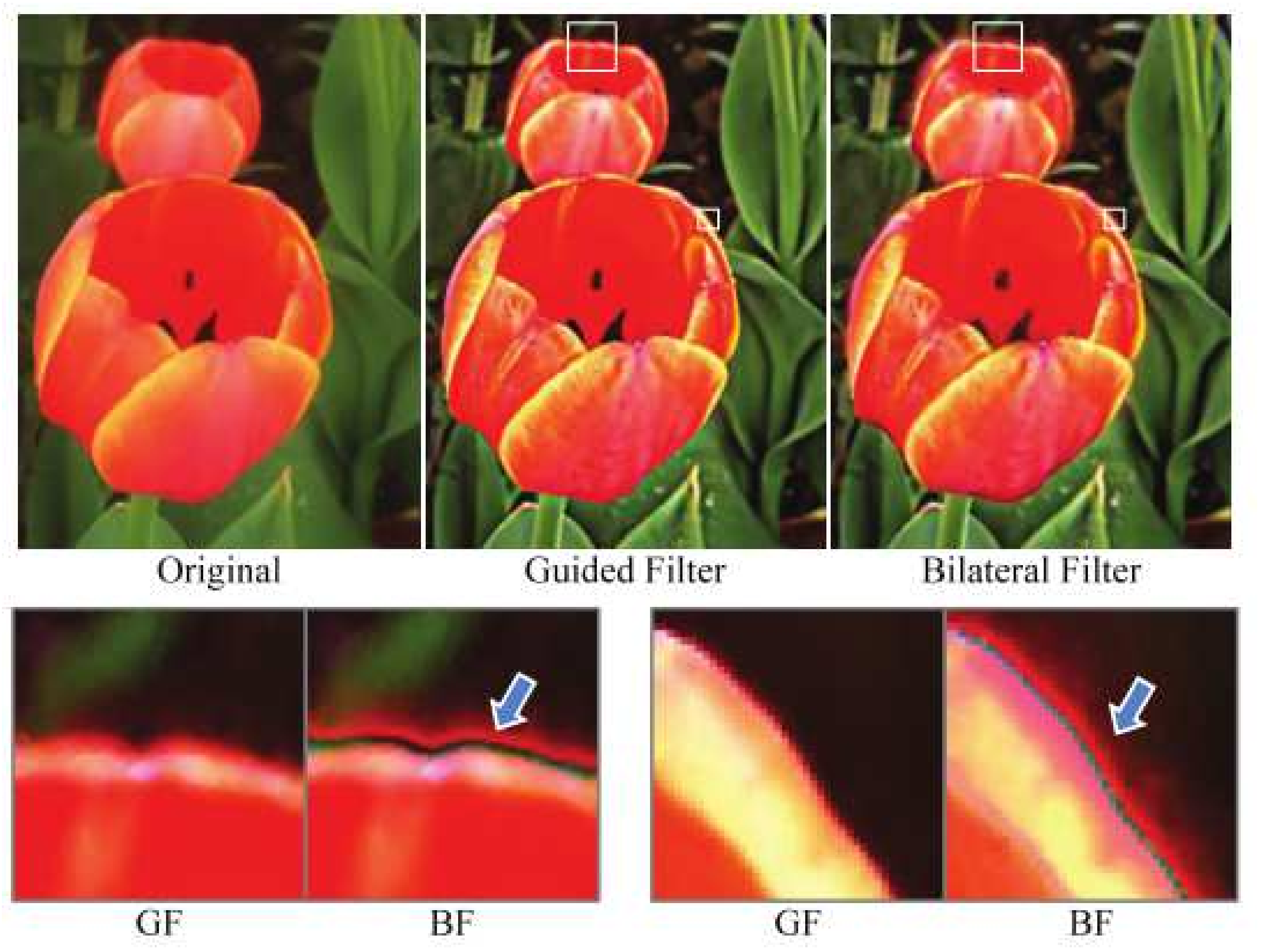}
\caption{Detail enhancement. The parameters are r=16, $\epsilon = 0.1^2$ for the guided filter, and $\sigma_s$=16, $\sigma_r$=0:1 for the bilateral filter. Figure reproduced from Kaiming He et al.\cite{kaiming}. BF= Bilateral filter, GF= Guided filter.}
\label{fig:fig39}
\end{figure}

\section{Conclusions and future scope}
In this review article we have tried to present the benchmark edge preserving smoothing algorithms present keeping the main focus on anisotropic diffusion and bilateral filtering. 
We have discussed the various efficient and variant ways (with less computational complexity and time), to implement the edge preserving filters together with the interrelationships with the structured numerical schemes with one another. We have discussed the various applications of the edge preserving filters keeping the focus on recent trends and their modifications and extensions in recent days as well, with an in-depth mathematical analysis. Our study on bilateral and anisotropic diffusion algorithm highlights several ways for future research. Moreover we have focused on hardware implementations of the most recent and efficient bilateral filter implementation and anisotropic diffusion filter with a view to make real time implementation making it a challenging task mainly in multidimensional data orientations used in computational photography. Future work will be to improve speed using parallel architecture (e.g GPU and using dedicated DSP blocks on reconfigurable architectures) to meet the strict timing constraints keeping in mind the trade-off with energy efficiency. Our in-depth study on the edge preserving filters in this paper will help the future research community of the computer vision domain to attain a detailed evolutionary understanding of the edge preservers, from heat diffusion equation to the most resent trilateral filters, their advantages and disadvantages and their existing interrelationship, to evolve with new innovative ideas more efficiently.

\section{Acknowledgement}

We thank Avik Kotal, Asit Samanta and Rourab Paul
for their encouragement and Dr.Kunal Narayan Choudhury for his advice. 
This work has been supported by the Department of Science and Technology, Govt of India under grant No
DST/INSPIRE FELLOWSHIP/2012/320 as well as the grant from TEQIP phase 2 ( COE ) of University of Calcutta
providing fund for this research.

\end{document}